\documentclass[twoside,11pt]{article}
\usepackage{jmlr2e}

\usepackage[utf8]{inputenc} 
\usepackage[T1]{fontenc}    
\usepackage{hyperref}       
\usepackage{url}            
\usepackage{booktabs}       
\usepackage{amsfonts}       
\usepackage{nicefrac}       
\usepackage{microtype}      
\usepackage{xcolor}         
\usepackage{amsmath}
\usepackage{amssymb}
\usepackage{graphicx}
\usepackage{subcaption}
\usepackage{array}
\usepackage{algorithm}
\usepackage{algorithmic}

\usepackage{amsmath}
\newtheorem{assumption}{Assumption}

\begin{document}

\title{Mean-Field Approximation of Cooperative Constrained Multi-Agent Reinforcement Learning (CMARL)}

\author{\name Washim Uddin Mondal \email wmondal@purdue.edu \\
	\addr Lyles School of Civil Engineering, \\
	School of Industrial Engineering,\\
	Purdue University, \\
	West Lafayette, IN, 47907, USA
	\AND
	\name Vaneet Aggarwal \email vaneet@purdue.edu \\
	\addr School of Industrial Engineering,\\
	School of Electrical and Computer Engineering,\\
	Purdue University, \\
	West Lafayette, IN, 47907, USA		
	\AND
	\name Satish V. Ukkusuri \email sukkusur@purdue.edu \\
	\addr Lyles School of Civil Engineering, \\
	Purdue University, \\
	West Lafayette, IN, 47907, USA
}

	\maketitle
	
	\begin{abstract}
		Mean-Field Control (MFC) has recently been proven to be a scalable tool to approximately solve large-scale multi-agent reinforcement learning (MARL) problems. However, these studies are typically limited to unconstrained cumulative reward maximization framework. In this paper, we show that one can use the MFC approach to approximate the MARL problem even in the presence of constraints. Specifically, we prove that, an $N$-agent constrained MARL problem, with state, and action spaces of each individual agents being of sizes $|\mathcal{X}|$, and $|\mathcal{U}|$ respectively, can be approximated by an associated constrained MFC problem with an  error, $e\triangleq \mathcal{O}\left([\sqrt{|\mathcal{X}|}+\sqrt{|\mathcal{U}|}]/\sqrt{N}\right)$. In a special case where the reward, cost, and state transition functions are independent of the action distribution of the population, we prove that the error can be improved to $e=\mathcal{O}(\sqrt{|\mathcal{X}|}/\sqrt{N})$. Also, we provide a Natural Policy Gradient based algorithm, and prove that it can solve the constrained MARL problem within an error of $\mathcal{O}(e)$ with a sample complexity of $\mathcal{O}(e^{-6})$.
	\end{abstract}
	
	\section{Introduction}
	\label{sec:intro}
	Consider the situation of a policy-maker that is tasked with allocating a certain amount of budget to improve the health of a number of infrastructures in a post-disaster scenario. If, due to the repair, the health of a certain infrastructure improves, it provides the perception of a high reward to the planner. However, such improvement comes at the ``cost'' of the amount of money that is allocated to it. How should the planner allocate the money over a long period of time? To answer such a question is the same as assigning a policy to each infrastructure that states how much money should that infrastructure draw from the budget pool based on the global state of all infrastructures. The aim of designing such policies should be to maximize  the  expected cumulative reward while obeying the  budget constraint.
	
	Strategizing the decisions of a large pool of interacting agents under certain constraints is a frequently appearing problem in many branches of social science and engineering. For example, such a framework can be applied to power-constrained wireless sensor networks \citep{buratti2009overview}, energy harvesting communication networks \citep{wang2015iterative},  queuing systems with stability constraints \citep{xiang2014joint}, etc. A common way to deal with such problems is to use the framework of constrained multi-agent reinforcement learning (CMARL). In cooperative CMARL, the goal is to maximize the aggregate reward of the whole population while obeying some specified constraints. Unfortunately, as the number of agents increases, the size of the joint state space increases exponentially, rendering the task of maximizing reward incredibly hard.
	
	The phenomenon of state-space explosion is not unique to CMARL, unconstrained MARL problems also have to deal with the same issue. There have been several attempts to alleviate this obstacle. For example, one idea is to confine oneself to local policies. This approach has yielded a number of useful algorithms e.g., Independent Q-Learning (IQL) \citep{tan1993multi}, centralized training and decentralized execution (CTDE) \citep{son2019qtran, sunehag2017value, rashid2020weighted, rashid2018qmix,zhou2022pac} based algorithms, etc. Despite their empirical success, none of the above algorithms can provide theoretical guarantees. Moreover, it is difficult to train policies using these algorithms when the number of agents is very high. Another alternative is the framework of mean-field control (MFC) which builds upon the idea that in an infinite population of identical agents, studying one representative agent is sufficient to infer the statistics of their collective behavior \citep{angiuli2022unified}. MFC has recently been applied to approximate $N$-agent unconstrained MARL problems with theoretical guarantees \citep{gu2021mean}. In contrast to the existing approaches, the benefit of the MFC is that its optimality gap decreases with an increase in the number of agents. It is, however,  still unknown whether similar approximation results can be established for CMARL problems. In this paper, we provide an affirmative answer to this question.
	
	Establishing MFC-based approximation bounds for CMARL is particularly challenging for the following reason. Unlike in unconstrained MARL, the set of feasible policies of CMARL may not be identical to the sets of feasible policies of its associated constrained MFC (CMFC) problem. In general, a feasible policy for CMARL may overshoot the constraints of CMFC and vice versa. Clearly, the sets of feasible policies of CMARL and CMFC may partially overlap but none is a subset of another. The lack of structural hierarchy between the feasible policy sets makes it difficult to compare the optimal values of CMARL and CMFC. The trick is not to deal with the original CMARL, and CMFC problems but to consider variants of those with smaller feasible sets. We show that these variants can be judiciously chosen to enforce hierarchy, and thereby create a way to compare values.

	
	\subsection{Our Contribution}
	\label{sec:our_contri}
	We consider an $N$-agent CMARL problem where at each instant the agents receive rewards as well as incur costs depending on their joint states, and actions. The goal is to maximize the time-discounted sum of rewards (called reward value or simple value) while ensuring that the discounted cumulative cost lies below a certain threshold. We show that the stated $N$-agent CMARL problem can be well approximated by a CMFC problem with an appropriately adjusted constraint. In particular, our result (Lemma \ref{theorem_1}) states that the optimal value of the stated CMFC is at most at the distance of $\mathcal{O}(e)$ from the optimal CMARL value where $e=\frac{1}{\sqrt{N}}[\sqrt{|\mathcal{X}|}+\sqrt{|\mathcal{U}|}]$. The terms $|\mathcal{X}|, |\mathcal{U}|$ denote the sizes of state, and action spaces of individual agents respectively. We also show that if the optimal policy obtained by solving the CMFC is adopted into the $N$-agent system, then it does not violate the constraint of CMARL, and yields an $N$-agent cumulative reward that is $\mathcal{O}(e)$ error away from the optimal $N$-agent value (Theorem \ref{corr_1}). In a special case where the reward, cost, and state transition functions are independent of the action distribution of the population, we prove that the error improves to, $e=\sqrt{|\mathcal{X}|}/\sqrt{N}$ (Theorem \ref{theorem_2}).

	The key idea behind Lemma \ref{theorem_1} is a novel sandwiching technique. Specifically, we consider three distinct problems$-$ a CMARL with zero constraint bound (target problem), a CMFC with slightly more restrictive constraint, and a CMARL with even more restrictive constraint. We prove that the optimal value for the CMFC problem must be trapped between the optimal values of the CMARL problems with a small margin of error. Next, we establish that the CMARL values themselves must be close for large $N$. This forces the CMFC value to lie in the vicinity of our target value.
	
	Finally, using the global convergence result of \citep{ding2020natural}, we devise a natural policy gradient-based primal-dual (NPG-PD) algorithm and show that the policy obtained from the stated algorithm satisfies the constraint of CMARL and approximates the optimal $N$-agent value within  an error of $\mathcal{O}(e)$. The sample complexity of the algorithm is shown to be $\mathcal{O}(e^{-6})$ (Theorem \ref{npg_theorem}).

	\subsection{Related Works}
	\label{sec:related_works}
	
	\textbf{Unconstrained Single Agent RL}: One of the early attempts to address the problem of single agent RL was made by the tabular Q-learning \citep{watkins1992q}, and subsequently by the SARSA algorithm \citep{rummery1994line}. However, such approaches work only when the state space is small. Neural Network-based Deep Q-learning \citep{mnih2015human}, and deep policy gradient algorithms \citep{li2019robust} have been introduced relatively recently to tackle large state space. One cannot, however, scale these approaches to a large number of agents due to the exponential blow-up of their joint state space.
	
	\textbf{Constrained Single Agent RL}: Decision making under constraint is typically modeled via Constrained Markov Decision Problems (CMDPs) \citep{chow2017risk}. Several policy gradient (PG) or direct policy search methods have been proposed to solve CMDPs \citep{achiam2017constrained, bhatnagar2012online}. However, the convergence guarantees of these algorithms are local. Recently, a series of PG-type algorithms have been proposed that ensure global convergence \citep{paternain2022safe, ding2020natural, mondal2024last}. The advantage of the mean-field approach is that it effectively reduces a multi-agent problem to a single-agent problem, thereby allowing us to utilize these existing guarantees of CMDPs.
	
	\textbf{MFC-based Approximation of MARL}: There have been a number of recent advances that show that MFC well-approximates MARL problems in the regime of a large population. \citep{gu2021mean} showed that an $N$-agent homogeneous MARL problem can be approximated by MFC within $\mathcal{O}(\frac{1}{\sqrt{N}})$ error. \citep{mondal2021approximation} later extended this idea to heterogeneous MARL. Their approaches, however, cannot be directly adopted for CMARL. MFC has also been used to design local policies \citep{mondal2022near} and approximate cooperative MARL with non-uniform interactions \citep{mondal2022can}.
	
	\textbf{Algorithms for MFC}: In the model-free set up, both $Q$-learning based \citep{angiuli2022unified, gu2021mean}, and PG-based algorithms \citep{carmona2019linear} have been proposed to solve MFC problems. Model-based algorithms are also available in the literature \citep{pasztor2021efficient}. However, none of these algorithms apply to CMFC problems.
	
	\textbf{Application of MFC}: MFC has found its application in a multitude of social and engineering applications ranging from epidemic control \citep{watkins2016optimal, lee2021controlling}, ride-hailing \citep{al2019deeppool}, network traffic engineering \citep{geng2020multi}, congestion management \citep{wang2020large} etc.
	
	\section{Model for Cooperative CMARL}
	\label{sec:cmarl}
	We consider a collection of $N$ agents interacting with each other in discrete time $t\in \{0,1,\cdots\}$. The state of $i$-th agent at time $t$ is denoted as $x_t^i$ which can take values from the finite state space, $\mathcal{X}$. The joint state of  $N$-agents at time $t$ is symbolised as $\boldsymbol{x}_t^N\triangleq\{x_t^i\}_{i\in \{1,\cdots, N\}}$. Upon observing $\boldsymbol{x}_t^N$, the $i$-th agent chooses an action $u_t^i$ from the finite action set, $\mathcal{U}$ according to the following probability law: $u_t^i\sim \pi_i(\boldsymbol{x}_t^N)$. The symbol, $\pi_i(\cdot)$, is defined to be the policy of the $i$-th agent which is assumed to be a function of the form $\pi_i:\mathcal{X}^N\rightarrow \mathcal{P}(\mathcal{U})$ where $\mathcal{P}(\cdot)$ defines a probability simplex over its argument. The joint action of $N$-agents at time $t$ is indicated as $\boldsymbol{u}_t^N\triangleq\{u_t^i\}_{i\in\{1,\cdots, N\}}$. As a consequence of these actions, the $i$-th agent receives a reward $r_i(\boldsymbol{x}_t^N, \boldsymbol{u}_t^N)$, a constraint cost $c_i(\boldsymbol{x}_t^N, \boldsymbol{u}_t^N)$ at time $t$, and its state changes in the next time step according to the following transition rule: $x_{t+1}^i\sim P_i(\boldsymbol{x}_t^N, \boldsymbol{u}_t^N)$. Here we implicitly assume that the elements of $\boldsymbol{u}_t^N$ are conditionally independent given $\boldsymbol{x}_t^N$  and the elements of $\boldsymbol{x}_{t+1}^N$ are conditionally independently given $(\boldsymbol{x}_t^N, \boldsymbol{u}_t^N)$.
	
	Note that, the reward, cost, and state transition law of an agent are not only affected by its own state, and action but also by states, and actions of other agents. Such interlacing is one of the main obstacles in solving large multi-agent problems. To ease the analysis, we shall assume a special structure of the reward, cost, and transition functions that are routinely adopted in the mean-field literature \citep{gu2021mean, angiuli2022unified}. Specifically, we assume that for some  $r, c: \mathcal{X}\times \mathcal{U} \times \mathcal{P}(\mathcal{X}) \times \mathcal{P}(\mathcal{U})\rightarrow \mathbb{R}$, and  $P: \mathcal{X}\times \mathcal{U} \times \mathcal{P}(\mathcal{X}) \times \mathcal{P}(\mathcal{U})\rightarrow \mathcal{P}(\mathcal{X})$, the following holds $\forall i\in \{1,\cdots, N\}$,
	\begin{align}
		\label{eq_1}
		&r_i(\boldsymbol{x}_t^N, \boldsymbol{u}_t^N) = r(x_t^i, u_t^i, \boldsymbol{\mu}_t^N, \boldsymbol{\nu}_t^N)\\
		\label{eq_2}
		&P_i(\boldsymbol{x}_t^N, \boldsymbol{u}_t^N) = P(x_t^i, u_t^i, \boldsymbol{\mu}_t^N, \boldsymbol{\nu}_t^N)\\
		\label{eq_3}
		&c_i(\boldsymbol{x}_t^N, \boldsymbol{u}_t^N) = c(x_t^i, u_t^i, \boldsymbol{\mu}_t^N, \boldsymbol{\nu}_t^N)
	\end{align}
	where $\boldsymbol{\mu}_t^N$, $\boldsymbol{\nu}_t^N$ denote the empirical $N$-agent state, and action distributions respectively. Formally,
	\begin{align}
	\label{eq_4}
		\boldsymbol{\mu}_t^N = \dfrac{1}{N}\sum_{i=1}^{N}\delta(x_t^i=x), ~\forall x\in \mathcal{X},~~\text{and}~~
		\boldsymbol{\nu}_t^N = \dfrac{1}{N}\sum_{i=1}^{N}\delta(u_t^i=u), ~\forall u\in \mathcal{U}
	\end{align}
	where $\delta(\cdot)$ is the indicator function. The equations $(\ref{eq_1})-(\ref{eq_3})$ are applicable to a system where agents are identical and exchangeable \citep{mondal2021approximation}. These relations suggests that the effect of the population on any agent is mediated only through the mean-field distributions. We would also like to point out that the functions, $r$, $P$, and $c$ are identical for all agents. Therefore, the system of agents described above is essentially homogeneous. As a consequence, we can rewrite the policy functions, $\pi_i$ as follows for some $\pi:\mathcal{X}\times \mathcal{P}(\mathcal{X})\rightarrow \mathcal{P}(\mathcal{U})$, and $\forall i\{1,\cdots, N\}$,
	\begin{align}
		\pi_i(\boldsymbol{x}_t^N) = \pi(x_t^i, \boldsymbol{\mu}_t^N)
	\end{align}
	
	In simple words, due to homogeneity, we can describe the policy of an agent as a (stochastic) process of choosing its action upon observing its own state, and the state distribution of the whole population. The policy of an agent at time $t$ is denoted as $\pi_t$ where its sequence is denoted as $\boldsymbol{\pi}\triangleq \{\pi_t\}_{t\in\{0,1,\cdots\}}$. For a given initial joint state, $\boldsymbol{x}^N_0$, the $N$-agent reward, and cost-value (indicated respectively as $V^{R}_{N}$, $V^{C}_{N}$) of a policy-sequence $\boldsymbol{\pi}$ are defined as follows.
	\begin{align}
		\label{eq_def_v_r}
		V^{R}_{N}(\boldsymbol{x}^N_0, \boldsymbol{\pi})\triangleq \sum_{t=0}^{\infty}\gamma^t \mathbb{E}\left[\dfrac{1}{N}\sum_{i=1}^N r(x_t^i, u_t^i, \boldsymbol{\mu}_t^N, \boldsymbol{\nu}_t^N)\right]\\
		\label{eq_def_v_c}
		V^{C}_{N}(\boldsymbol{x}^N_0, \boldsymbol{\pi})\triangleq \sum_{t=0}^{\infty}\gamma^t \mathbb{E}\left[\dfrac{1}{N}\sum_{i=1}^N c(x_t^i, u_t^i, \boldsymbol{\mu}_t^N, \boldsymbol{\nu}_t^N)\right]
	\end{align}
	where the expectation is computed over all joint state-action trajectories induced by $\boldsymbol{\pi}$, and $\gamma\in[0, 1)$ is a discount factor. Let the set of all admissible policies be $\Pi$, and $\Pi_{\infty}\triangleq \Pi \times \Pi \times \cdots$ be the collection of all admissible policy sequences. The goal of $N$-agent CMARL is to solve the following optimization problem for a given initial joint state $\boldsymbol{x}_0^N$.
	\begin{align}
		\tag{CMARL}
		\label{problem_1}
		\begin{split}
			\sup_{\boldsymbol{\pi}\in \Pi_{\infty}}&~V_{N}^{R}(\boldsymbol{x}^N_0, \boldsymbol{\pi})~~~~
			\mathrm{subject~to}: ~V_{N}^{C}(\boldsymbol{x}^N_0, \boldsymbol{\pi}) \leq 0
		\end{split}
	\end{align}
	
	We shall denote the solution to (\ref{problem_1}) as $\boldsymbol{\pi}^*_{N}$. The existence of an optimal solution is guaranteed via Slater's condition described later. Obtaining a solution of (\ref{problem_1}), is, in general, difficult. In the next section, we shall demonstrate how $\boldsymbol{\pi}^*_{N}$ can be approximated via the constrained mean field control (CMFC) approach.
	
	\section{Model for Constrained Mean-Field Control (CMFC)}
	\label{sec:cmfc}
	
	A mean-field system comprises an infinite collection of agents. Due to homogeneity, it is sufficient to track only a representative agent to describe the behavior of the population. Let the state, and action of this representative at time $t$ be denoted as $x_t\in\mathcal{X}$, $u_t\in\mathcal{U}$ respectively while the distributions of states, and actions of the whole population be denoted as $\boldsymbol{\mu}_t^\infty\in\mathcal{P}(\mathcal{X})$, and $\boldsymbol{\nu}_t^\infty\in\mathcal{P}(\mathcal{U})$. Observe that, for a given policy sequence, $\boldsymbol{\pi}\triangleq \{\pi_t\}_{t\in\{0,1,\cdots\}}$, the action distribution, $\boldsymbol{\nu}_t^\infty$ can be derived from the state distribution, $\boldsymbol{\mu}_t^\infty$ as follows.
	\begin{align}
		\label{eq:nu_t}
		\boldsymbol{\nu}_t^\infty = \nu^{\mathrm{MF}}(\boldsymbol{\mu}_t^\infty, \pi_t) \triangleq \sum_{x\in\mathcal{X}}\pi_t(x, \boldsymbol{\mu}_t^\infty)\boldsymbol{\mu}_t^\infty(x)
	\end{align}
	
	Similarly, the state distribution at time $t+1$ can be obtained from $\boldsymbol{\mu}_t^\infty$ using the following update equation.
	\begin{align}
		\label{eq:mu_t_1}
		\boldsymbol{\mu}_{t+1}^\infty = P^{\mathrm{MF}}(\boldsymbol{\mu}_t^\infty, \pi_t) \triangleq \sum_{x\in\mathcal{X}}\sum_{u\in\mathcal{U}} P(x, u, \boldsymbol{\mu}_t^\infty, \nu^{\mathrm{MF}}(\boldsymbol{\mu}_t^\infty, \pi_t)) \pi_t(x, \boldsymbol{\mu}_t^\infty)(u) \boldsymbol{\mu}_t^\infty(x) 
	\end{align}
	
	Finally, the average reward, and constraint cost at time $t$ can be computed as
	\begin{align}
		\label{eq:r_MF}
		r^{\mathrm{MF}}(\boldsymbol{\mu}_t^\infty, \pi_t) = \sum_{x\in\mathcal{X}}\sum_{u\in\mathcal{U}} r(x, u, \boldsymbol{\mu}_t^\infty, \nu^{\mathrm{MF}}(\boldsymbol{\mu}_t^\infty, \pi_t)) \pi_t(x, \boldsymbol{\mu}_t^\infty)(u)\boldsymbol{\mu}_t^\infty(x)\\
		\label{eq:c_MF}
		c^{\mathrm{MF}}(\boldsymbol{\mu}_t^\infty, \pi_t) = \sum_{x\in\mathcal{X}}\sum_{u\in\mathcal{U}} c(x, u, \boldsymbol{\mu}_t^\infty, \nu^{\mathrm{MF}}(\boldsymbol{\mu}_t^\infty, \pi_t)) \pi_t(x, \boldsymbol{\mu}_t^\infty)(u)\boldsymbol{\mu}_t^\infty(x)
	\end{align}
	
	For an initial state distribution, $\boldsymbol{\mu}_0^\infty = \boldsymbol{\mu}_0$, and a policy-sequence $\boldsymbol{\pi}$, we define the value functions related to reward, and constraint cost as follows.
	\begin{align}
		\label{eq:v_r_mf}
		V^{R}_{\infty}(\boldsymbol{\mu}_0, \boldsymbol{\pi}) = \sum_{t=0}^\infty \gamma^t r^{\mathrm{MF}}(\boldsymbol{\mu}_t^\infty, \pi_t)\\
		\label{eq:v_c_mf}
		V^{C}_{\infty}(\boldsymbol{\mu}_0, \boldsymbol{\pi}) = \sum_{t=0}^\infty \gamma^t c^{\mathrm{MF}}(\boldsymbol{\mu}_t^\infty, \pi_t)
	\end{align}
	
	The goal of CMFC is to solve the following optimization problem.
	\begin{align}
		\label{eq:problem_2}
		\tag{CMFC}
		\begin{split}
			\sup_{\boldsymbol{\pi}\in \Pi_{\infty}}&~V_{\infty}^{R}(\boldsymbol{\mu}_0, \boldsymbol{\pi})~~~~
			\mathrm{subject~to}: ~V_{\infty}^{C}(\boldsymbol{\mu}_0, \boldsymbol{\pi}) \leq 0
		\end{split}
	\end{align}
	where $\Pi_{\infty}$, as stated in section \ref{sec:cmarl}, is the set of admissible policy sequences. In the forthcoming section, we shall establish that the solution of a variant of (\ref{eq:problem_2}) closely approximates the solution of (\ref{problem_1}) for large $N$.
	
	\section{Main Result: Approximation of CMARL via CMFC}
	\label{sec:approximation_cmarl_via_cmfc}
	
	Before stating the approximation result, we would like to enlist the set of assumptions needed to prove it. Our first assumption is on the reward, cost, and state transition function.
	
	\begin{assumption}
		\label{assumption_1}
		There exists constants $M_R, M_C, L_R, L_C, L_P>0$ such that the following relations hold $\forall x\in \mathcal{X}$, $\forall u\in \mathcal{U}$, $\forall \boldsymbol{\mu}_1, \boldsymbol{\mu}_2\in \mathcal{P}(\mathcal{X})$, and $\forall \boldsymbol{\nu}_1, \boldsymbol{\nu}_2\in \mathcal{P}(\mathcal{U})$.
		\begin{align*}
			&(a) |r(x, u, \boldsymbol{\mu}_1, \boldsymbol{\nu}_1)| \leq M_R,\\
			&(b) |c(x, u, \boldsymbol{\mu}_1, \boldsymbol{\nu}_1)| \leq M_C,\\
			&(c) |r(x, u, \boldsymbol{\mu}_1, \boldsymbol{\nu}_1) - r(x, u, \boldsymbol{\mu}_2, \boldsymbol{\nu}_2)| \leq L_R\left[|\boldsymbol{\mu}_1-\boldsymbol{\mu}_2|_1+|\boldsymbol{\nu}_1-\boldsymbol{\nu}_2|_1\right],\\
			&(d) |c(x, u, \boldsymbol{\mu}_1, \boldsymbol{\nu}_1) - c(x, u, \boldsymbol{\mu}_2, \boldsymbol{\nu}_2)| \leq L_C\left[|\boldsymbol{\mu}_1-\boldsymbol{\mu}_2|_1+|\boldsymbol{\nu}_1-\boldsymbol{\nu}_2|_1\right],\\
			&(e) |P(x, u, \boldsymbol{\mu}_1, \boldsymbol{\nu}_1) - P(x, u, \boldsymbol{\mu}_2, \boldsymbol{\nu}_2)|_1 \leq L_P\left[|\boldsymbol{\mu}_1-\boldsymbol{\mu}_2|_1+|\boldsymbol{\nu}_1-\boldsymbol{\nu}_2|_1\right]
		\end{align*} 
		where $|\cdot|_1$ denotes $L_1$-distance. 
	\end{assumption} 
	
	Assumption \ref{assumption_1}(a), and \ref{assumption_1}(b) states that the reward, $r$, and the cost function, $c$ are bounded. The very definition of the transition function, $P$ makes it bounded. Hence, it is not listed as an assumption. On the other hand, Assumption \ref{assumption_1}(c)-(e) dictate that the functions $r$, $c$, and $P$ are Lipschitz continuous with respect to their state, and action distribution arguments. Assumption \ref{assumption_1} is common in the  literature \citep{angiuli2022unified, carmona2018probabilistic}. Our next assumption is on the class of admissible policies, $\Pi$.
	\begin{assumption}
		\label{assumption_policy}
		There exists a constant $L_Q>0$ such that $\forall \pi\in\Pi$ the following holds.
		\begin{align*}
			|\pi(x, \boldsymbol{\mu}_1) - \pi(x, \boldsymbol{\mu}_2)|\leq L_Q|\boldsymbol{\mu}_1-\boldsymbol{\mu}_2|_1
		\end{align*}
		$\forall x\in \mathcal{X}$, and $\forall \boldsymbol{\mu}_1, \boldsymbol{\mu}_2\in \mathcal{P}(\mathcal{X})$.
	\end{assumption}
	
	Assumption \ref{assumption_policy} dictates that the admissible policies are Lipschitz continuous with respect to the state distributions. The assumption stated above is also common in the literature and holds for neural network-based policies with bounded weights \citep{gu2021mean, pasztor2021efficient}.
	Before stating the main result, we need to introduce some notations. Consider the following optimization problem.
	\begin{align}
		\tag{G-CMARL}
		\label{problem_1a}
		\begin{split}
			\sup_{\boldsymbol{\pi}\in \Pi_{\infty}}&~V_{N}^{R}(\boldsymbol{x}^N_0, \boldsymbol{\pi})~~~~
			\mathrm{subject~to}: ~V_{N}^{C}(\boldsymbol{x}^N_0, \boldsymbol{\pi}) \leq \zeta
		\end{split}
	\end{align}
	
	Problem (\ref{problem_1a}) is a generalization of the problem (\ref{problem_1}). Specifically, it uses an arbitrary real, $\zeta\in \mathbb{R}$, as the constraint upper bound, instead of $\zeta=0$ as considered in (\ref{problem_1}). With slight abuse of notation, we define the optimal objective value of (\ref{problem_1a}) as $V_N^*(\boldsymbol{x}_0^N, \zeta)$. In a similar fashion, $V_{\infty}^*(\boldsymbol{\mu}_0, \zeta)$ denotes the optimal objective value of the following  optimization problem for $\zeta\in \mathbb{R}$.
	\begin{align}
		\tag{G-CMFC}
		\label{problem_2a}
		\begin{split}
			\sup_{\boldsymbol{\pi}\in \Pi_{\infty}}&~V_{\infty}^{R}(\boldsymbol{\mu}_0, \boldsymbol{\pi})~~~~
			\mathrm{subject~to}:~ ~V_{\infty}^{C}(\boldsymbol{\mu}_0, \boldsymbol{\pi}) \leq \zeta
		\end{split}
	\end{align}

        \begin{table}[]
        \centering
        \begin{tabular}{|c|p{7.8cm}|c|}
            \hline
             Notation & Interpretation & Used In  \\
             \hline
             $\boldsymbol{\pi}_N^*$ & Optimal solution of (\ref{problem_1}) & Theorem \ref{corr_1}\\
             \hline
            $\boldsymbol{\pi}_0^*$ & Optimal solution of (\ref{problem_1}) & Section \ref{sec:proof_theorem_1}\\
            \hline
             $\boldsymbol{\pi}_{\infty}^*$ & Optimal solution of (\ref{problem_2a}) for $\zeta=-G_C$ & Theorem \ref{corr_1}, \ref{theorem_2}\\
            \hline
             $\boldsymbol{\pi}_1^*$  & Optimal solution of (\ref{problem_2a}) for $\zeta=-G_C$ & Section \ref{sec:proof_theorem_1}\\
            \hline
             $\boldsymbol{\pi}_2^*$  & Optimal solution of (\ref{problem_1a}) for $\zeta=-2G_C$ & Section \ref{sec:proof_theorem_1}\\
             \hline
             $V_N^R(\boldsymbol{x}_0^N, \boldsymbol{\pi})$, & Reward, constraint values of $N$-agent system    & Theorem \ref{corr_1}, \ref{theorem_2}, \ref{npg_theorem}, \\   $V_N^C(\boldsymbol{x}_0^N, \boldsymbol{\pi})$ & for initial states $\boldsymbol{x}_0^N$, and policy sequence $\boldsymbol{\pi}$  & Lemma \ref{lemma_1a}\\
            \hline
                 $V_{\infty}^R(\boldsymbol{\mu}_0, \boldsymbol{\pi})$, & Reward, constraint values of MFC system for   & Theorem \ref{corr_1}, \ref{theorem_2}, \ref{npg_theorem}, \\   $V_{\infty}^C(\boldsymbol{\mu}_0, \boldsymbol{\pi})$ & initial distribution $\boldsymbol{\mu}_0$, and policy sequence $\boldsymbol{\pi}$  & Lemma \ref{lemma_1a}\\
             \hline
             $V_N^*(\boldsymbol{x}_0^N, \zeta)$, & Optimal objective values of (\ref{problem_1a}) and  & Lemma \ref{theorem_1}, Theorem \ref{theorem_2}, \ref{npg_theorem}\\
             $V_{\infty}^*(\boldsymbol{x}_0^N, \zeta)$ & (\ref{problem_2a}) respectively with constraint $\zeta$ & Section \ref{sec:proof_theorem_1}\\
             \hline
        \end{tabular}
        \caption{Notations. Note that $V_N^*(\boldsymbol{x}_0^N, -2G_C)=V_N^R(\boldsymbol{x}_0^N, \boldsymbol{\pi}_2^*)$, $V_{\infty}^*(\boldsymbol{\mu}_0, -G_C) = V_{\infty}^R(\boldsymbol{\mu}_0, \boldsymbol{\pi}_1^*)$ $=V_{\infty}^R(\boldsymbol{\mu}_0, \boldsymbol{\pi}_{\infty}^*)$, and $V_N^*(\boldsymbol{x}_0^N, 0)=V_N^R(\boldsymbol{x}_0^N, \boldsymbol{\pi}_0^*)=V_N^R(\boldsymbol{x}_0^N, \boldsymbol{\pi}_N^*)$.}
        \label{table_notation}
    \end{table}

	Clearly, (\ref{problem_2a}) generalizes the problem, (\ref{eq:problem_2}). Table \ref{table_notation} summarizes the notations used in the paper. The following Assumption is required (in addition to Assumption \ref{assumption_1}, \ref{assumption_policy}) to establish the main result. 
	\begin{assumption}
		\label{assumption_2}
		There exists $\zeta_0>0$ such that the set of feasible solutions of (\ref{problem_1a}) for $\zeta = -\zeta_0$ is non-empty. In other words, there exists a policy-sequence, $\boldsymbol{\pi}\in \Pi_{\infty}$ such that $V_N^C(\boldsymbol{x}_0^N, \boldsymbol{\pi})\leq -\zeta_0$ for any given joint initial state, $\boldsymbol{x}_0^N$.
	\end{assumption}
	
	Assumption \ref{assumption_2} ensures that no pathological example with an empty feasible set arises in our analysis. This assumption is similar to Slater's condition in the optimization literature. Below we state an intermediate result that will be useful to prove our main theorem.

	\begin{lemma}
		\label{theorem_1}
		Let $\boldsymbol{x}_0^N$ denote the initial joint state in an $N$-agent system, and $\boldsymbol{\mu}_0$ be its empirical distribution. If assumptions $\ref{assumption_1}-\ref{assumption_2}$ hold, then there exists a sufficiently large $N_0>0$ such that $\forall N\geq N_0$ the following inequality holds whenever $\gamma S_P<1$.
		\begin{align}
			\label{eq_15}
			|V_N^*(\boldsymbol{x}_0^N, 0) - V_{\infty}^*(\boldsymbol{\mu}_0, -G_C)| \leq G_R + G_C\left[\dfrac{4}{\zeta_0}\left(\dfrac{M_R}{1-\gamma}\right)\right]
		\end{align}
		The terms $G_R, G_C$ are defined as shown below. 
		\begin{align}
			\label{def_G_J}
			G_J\triangleq \left(\dfrac{1}{1-\gamma}\right)\left[\dfrac{M_J}{\sqrt{N}}+\dfrac{L_J\sqrt{|\mathcal{U}|}}{\sqrt{N}}\right] 
			+\dfrac{\left[\sqrt{|\mathcal{X}|}+\sqrt{|\mathcal{U}|}\right]}{\sqrt{N}} \left(\dfrac{ S_JC_P}{S_P-1}\right)\left[\dfrac{1}{1-\gamma S_P}-\dfrac{1}{1-\gamma}\right]
		\end{align}
		where $C_P\triangleq 2+L_P$, $S_J\triangleq (M_J+2L_J) + L_Q(M_J+L_J)$, $S_P\triangleq (1+2L_J) + L_Q(1+L_J)$, and $J\in \{R, C\}$. The terms $\{M_R, M_C, L_R, L_C, L_P, \zeta_0\}$ are defined in Assumptions \ref{assumption_1},  \ref{assumption_policy}, and \ref{assumption_2}.
	\end{lemma}
	
	Lemma \ref{theorem_1} provides a recipe for approximately solving the $N$-agent constrained MARL problem (\ref{problem_1}).  In particular,
	inequality $(\ref{eq_15})$ ensures that the optimal value function of $N$-agent problem (\ref{problem_1}), and that of generalized constrained mean-field problem (\ref{problem_2a}) for $\zeta = -G_C$ are at most $\mathcal{O}(G_R+G_C)$ distance away from each other. Note that $G_R, G_C=\mathcal{O}\left(\frac{1}{\sqrt{N}}\right)$. Thus for sufficiently large $N$, obtaining a solution for (\ref{problem_2a}) for $\zeta=-G_C$ is approximately equivalent to solving the original $N$-agent problem (\ref{problem_1}). We would like to mention that, $G_R, G_C = \mathcal{O}(\sqrt{|\mathcal{X}|}+\sqrt{|\mathcal{U}|})$ where $|\mathcal{X}|, |\mathcal{U}|$ denote the sizes of state, and action spaces respectively. Therefore, as the state and action spaces of individual agents become larger, the approximation progressive becomes worse.

	Although $(\ref{eq_15})$ establishes the proximity of the value-functions of $N$-agent, and infinite agent systems, it does not clarify whether the policy sequence obtained by solving (\ref{problem_2a}) for $\zeta = -G_C$ would obey the constraint of (\ref{problem_1}) if the policy is adopted in an $N$-agent system. It also does not comment on how the said policy sequence would perform (in terms of cumulative reward) in an $N$-agent system in comparison to the optimal $N$-agent value. These important questions are answered in the following Theorem. The proof of Theorem \ref{corr_1} is relegated to Appendix \ref{sec:appndx_corr_1}.
	
	\begin{theorem}
		\label{corr_1}
		Let $\boldsymbol{\pi}^*_{N}$, $\boldsymbol{\pi}^*_{\infty}$ be solutions of (\ref{problem_1}), and (\ref{problem_2a}) with $\zeta= -G_C$ respectively. If Assumptions $\ref{assumption_1}- \ref{assumption_2}$ hold, then the following inequality holds for $\forall N\geq N_0$ whenever one has $\gamma S_P<1$.
		\begin{align}
			\label{eq_17}
			&V_N^{R}(\boldsymbol{x}_0^N, \boldsymbol{\pi}_N^*) - V_N^{R}(\boldsymbol{x}_0^N, \boldsymbol{\pi}^*_{\infty}) \leq  2G_R + G_C\left[\dfrac{4}{\zeta_0}\left(\dfrac{M_R}{1-\gamma}\right)\right],\\
			\label{eq_18}
			\text{and }& V_N^{C}(\boldsymbol{x}_0^N, \boldsymbol{\pi}_{\infty}^*) \leq 0
		\end{align}
		where $V_N^{R}(\cdot, \cdot)$, and $V_N^{C}(\cdot, \cdot)$ are defined in $(\ref{eq_def_v_r})$, and $(\ref{eq_def_v_c})$ respectively. The notations $\boldsymbol{x}_0^N$, $\boldsymbol{\mu}_0$, $G_{R}$, $G_{C}$, $S_P$, and $N_0$ carry the same meaning as stated in Lemma \ref{theorem_1}.
	\end{theorem}

	Inequalities $(\ref{eq_17}), (\ref{eq_18})$ dictate that if the policy-sequence, $\boldsymbol{\pi}^*_{\infty}$, obtained by solving the mean-field problem (\ref{problem_2a}) for $\zeta = -G_C$ is adopted in the $N$-agent system, then the  value obtained for such a system is $\mathcal{O}(\frac{1}{\sqrt{N}})$ error away from the optimum. Also, $\boldsymbol{\pi}_{\infty}^*$ does not violate (\ref{problem_1}) constraint.

	\section{Proof of Lemma \ref{theorem_1}}
	\label{sec:proof_theorem_1}
	
	The proof of Lemma \ref{theorem_1} hinges on Lemma \ref{lemma_1a}. The proof of Lemma \ref{lemma_1a} is given in Appendix \ref{sec:appndx_lemma_1a}.
	\begin{lemma}
		\label{lemma_1a}
		Let $\boldsymbol{x}_0^N$ denote the initial joint state in an $N$-agent system, and $\boldsymbol{\mu}_0$ be its empirical distribution. If Assumption $\ref{assumption_1}-\ref{assumption_2}$ hold, then there exists sufficiently large $N_0>0$ such that $\forall N\geq N_0$, and $\forall \boldsymbol{\pi}\in\Pi_{\infty}$, the following inequalities hold whenever $\gamma S_P<1$.
		\begin{align}
			\label{eq_19}
			&|V_N^J(\boldsymbol{x}_0^N, \boldsymbol{\pi}) - V_{\infty}^J(\boldsymbol{\mu}_0, \boldsymbol{\pi})|\leq G_J
		\end{align}
		where $J\in \{R, C\}$. The value functions $V_N^{R}, V_N^{C}, V_{\infty}^R, V_\infty^C$ are provided by $(\ref{eq_def_v_r}), (\ref{eq_def_v_c})$, $(\ref{eq:v_r_mf}), (\ref{eq:v_c_mf})$ respectively, and the terms $G_{R}, G_{C}, S_P$ are defined in Lemma \ref{theorem_1}.
	\end{lemma}
	
	Intuitively, Lemma \ref{lemma_1a} dictates that the value function pairs $(V_N^{R}, V_{\infty}^{R})$, $(V_N^{C}, V_{\infty}^{C})$ are close in the sense that for every policy-sequence $\boldsymbol{\pi}$, the differences $|V_N^J(\boldsymbol{x}_0^N, \boldsymbol{\pi}) - V_{\infty}^J(\boldsymbol{\mu}_0, \boldsymbol{\pi})|, |V_N^J(\boldsymbol{x}_0^N, \boldsymbol{\pi}) - V_{\infty}^J(\boldsymbol{\mu}_0, \boldsymbol{\pi})|$, $J\in\{R, C\}$ are of the form $\mathcal{O}(\frac{1}{\sqrt{N}})$. Below we explain how such a result helps us to establish Lemma \ref{theorem_1}. Consider the three distinct constrained maximizations stated below.
	\begin{itemize}
		\item Problem $0$: (\ref{problem_1a}) with $\zeta = 0$. It is the same as the original $N$-agent problem (\ref{problem_1}). 
		\item Problem $1$: (\ref{problem_2a}) with $\zeta = -G_{C}$.
		\item Problem $2$: (\ref{problem_1a}) with $\zeta = -2G_{C}$.
	\end{itemize}
	
	Let, $\Pi_{\infty}^k\subset \Pi_{\infty}$ be the set of admissible policy sequences that satisfy the constraint of $k$-th optimization problem stated above, $k\in\{0,1,2\}$. Note that, if $\boldsymbol{\pi}\in \Pi_{\infty}^2$, then
	\begin{align}
		V_{\infty}^{C}(\boldsymbol{\mu}_0, \boldsymbol{\pi})\overset{(a)}{\leq} V_N^{C}(\boldsymbol{x}_0^N, \boldsymbol{\pi}) + G_C \overset{(b)}{\leq} -G_C
	\end{align}
	
	Inequality $(a)$ follows from $(\ref{eq_19})$ and $(b)$ is a consequence of the fact that $\boldsymbol{\pi}\in \Pi_{\infty}^2$. Thus, $\Pi_{\infty}^2\subset \Pi_{\infty}^1$. Similarly, if $\boldsymbol{\pi}\in \Pi_{\infty}^1$, then
	\begin{align}
		V_N^{C}(\boldsymbol{x}_0^N, \boldsymbol{\pi}) \overset{(a)}{\leq} V_{\infty}^{C}(\boldsymbol{\mu}_0, \boldsymbol{\pi})  + G_C \overset{(b)}{\leq} 0
	\end{align}
	which shows $\Pi_{\infty}^1\subset \Pi_{\infty}^0$. Combining, we get $\Pi_{\infty}^2\subset \Pi_{\infty}^1 \subset \Pi_{\infty}^0$. Finally, assume that $\boldsymbol{\pi}^*_k\in \Pi_{\infty}^k$ indicate the solution of $k$-th maximization problem described above, $k\in\{0,1,2\}$. Then, 
	\begin{align}
		\label{eq22}
		&	V_{\infty}^{R}(\boldsymbol{\mu}_0, \boldsymbol{\pi}_1^*)\overset{(a)}{\leq} V_N^{R}(\boldsymbol{x}_0^N, \boldsymbol{\pi}^*_1)+G_R\overset{(b)}{\leq} 	V_N^{R}(\boldsymbol{x}_0^N, \boldsymbol{\pi}^*_0)+G_{R} = V_N^*(\boldsymbol{x}_0^N, 0) + G_{R},\\
		\label{eq23}
		&	V_{\infty}^{R}(\boldsymbol{\mu}_0, \boldsymbol{\pi}_1^*)\overset{(c)}{\geq}V_{\infty}^{R}(\boldsymbol{\mu}_0, \boldsymbol{\pi}_2^*) \overset{(d)}{\geq} V_N^{R}(\boldsymbol{x}_0^N, \boldsymbol{\pi}^*_2)-G_{R}= V_N^*(\boldsymbol{x}_0^N, -2G_{C}) - G_{R}
	\end{align}
	
	Inequalities (a), and (d) follow from $(\ref{eq_19})$. Relation $(b)$ is a consequence of the fact that $\boldsymbol{\pi}_0^*$ is a maximizer of $V_N^{R}(\boldsymbol{x}_0^N, \cdot)$ over $\Pi_{\infty}^0$, and $\boldsymbol{\pi}_1^*\in \Pi_{\infty}^1\subset \Pi_{\infty}^0$. Similarly, $\boldsymbol{\pi}_1^*$ is a maximizer of $V_{\infty}^{R}(\boldsymbol{\mu}_0, \cdot)$ over $\Pi_{\infty}^1$, and $\boldsymbol{\pi}_2^*\in \Pi_{\infty}^2\subset \Pi_{\infty}^1$. This establishes relation (c).
	
	The gist of the above exercise is that the value $V_{\infty}^*(\boldsymbol{\mu}_0, -G_C)$ is sandwiched between $V_N^*(\boldsymbol{x}_0^N, 0)+G_R$, and $V_N^*(\boldsymbol{x}_0^N, -2G_C)-G_R$. To prove $(\ref{eq_15})$, one needs to bound the distance between $V_N^*(\boldsymbol{x}_0^N, 0)$, and $V_N^*(\boldsymbol{x}_0^N, -2G_C)$. This can be accomplished utilizing the concavity property of $V_N^*(\boldsymbol{x}_0^N, \cdot)$ in the domain $[-\zeta_0, 0]$ where $\zeta_0$ is given in Assumption \ref{assumption_2}. The concavity property is a consequence of Proposition 1 of \citep{paternain2019constrained}.  If $N\geq N_0$ where $N_0$ is a sufficiently large value, then $2G_C<\zeta_0$. In such a case,
	\begin{align*}
		&V_N^*(\boldsymbol{x}_0^N, -2G_C) \geq \left(1-\dfrac{2G_C}{\zeta_0}\right)V_N^*(\boldsymbol{x},0)+ \dfrac{2G_C}{\zeta_0} 	V_N^*(\boldsymbol{x}_0^N,-\zeta_0)\\
		\text{Equivalently, }& V_N^*(\boldsymbol{x}_0^N,0) - V_N^*(\boldsymbol{x}, -2G_C) \leq \dfrac{2G_C}{\zeta_0}\left[V_N^*(\boldsymbol{x}_0^N, 0)-V_N^*(\boldsymbol{x}_0^N, -\zeta_0)\right]
	\end{align*}
	
	Using Assumption \ref{assumption_1}(a), one can trivially deduce, $|V_N^*(\boldsymbol{x}_0^N, \zeta)|\leq M_R/(1-\gamma)$, $\forall \zeta \geq -\zeta_0$. More specifically, this is true for $\zeta=-\zeta_0, 0$. Hence,
	\begin{align}
		\label{eq_22}
		V_N^*(\boldsymbol{x}_0^N,0) - V_N^*(\boldsymbol{x}_0^N, -2G_C) \leq G_C\left[\dfrac{4}{\zeta_0}\left(\dfrac{M_R}{1-\gamma}\right)\right]
	\end{align}
	
	Using $(\ref{eq23}), (\ref{eq_22})$, we obtain,
	\begin{align}
		\label{eq_25}
		V_N^*(\boldsymbol{x}_0^N,0) - V_{\infty}^*(\boldsymbol{\mu}_0, -G_C) \leq G_R+G_C\left[\dfrac{4}{\zeta_0}\left(\dfrac{M_R}{1-\gamma}\right)\right]
	\end{align}
	
	Combining with $(\ref{eq22})$, we conclude the result.

	\section{Improvement of Optimality Gap in a Special Case}
	\label{sec:improvement}
	
	In this section, we shall impose the following additional restriction on the structure of reward, $r$, cost, $c$, and transition function, $P$ to improve the approximation bound of Lemma \ref{theorem_1}.
	
	\begin{assumption}
		\label{assumption_3}
		The functions $r$, $c$, and $P$ are independent of the action distribution of the population. Mathematically, $\forall x\in \mathcal{X}$, $\forall u\in \mathcal{U}$, $\forall \boldsymbol{\mu}\in \mathcal{P}(\mathcal{X})$, and $\forall \boldsymbol{\nu}\in \mathcal{P}(\mathcal{U})$, 
		\begin{align*}
			&	(a) ~r(x, u, \boldsymbol{\mu}, \boldsymbol{\nu}) = r(x, u, \boldsymbol{\mu}),
			(b)~c(x, u, \boldsymbol{\mu}, \boldsymbol{\nu}) = c(x, u, \boldsymbol{\mu}),~
			(c)~P(x, u, \boldsymbol{\mu}, \boldsymbol{\nu}) = P(x, u, \boldsymbol{\mu})
		\end{align*}
	\end{assumption}
	
	Assumption \ref{assumption_3} removes the dependence of $r, c$, and $P$ on the action distribution. However, for each agent, those functions still take the action executed by the same agent as an argument. Below we present our improved approximation result.
	\begin{theorem}
		\label{theorem_2}
		Let $\boldsymbol{x}_0^N$ denote an initial joint state in an $N$-agent system, and $\boldsymbol{\mu}_0$ its empirical distribution. If Assumptions $\ref{assumption_1}- \ref{assumption_3}$ are true, then there exists a sufficiently large $N_0>0$ such that $\forall N\geq N_0$ the following inequality holds whenever $\gamma S_P<1$.
		\begin{align}
			\label{eq_threorem_2_1}
			|V_N^*(\boldsymbol{x}_0^N, 0) - V_{\infty}^*(\boldsymbol{\mu}_0, -G_C^0)| \leq G_R^0 + G_C^0\left[\dfrac{4}{\zeta_0}\left(\dfrac{M_R}{1-\gamma}\right)\right]
		\end{align}
		The terms $G_R^0, G_C^0$ are defined as shown below. 
		\begin{align}
			\label{def_G_J_thm_2}
			G_J^0\triangleq \dfrac{M_J}{\sqrt{N}}\left(\dfrac{1}{1-\gamma}\right) 
			+\dfrac{\sqrt{|\mathcal{X}|}}{\sqrt{N}} \left(\dfrac{ 2S_J}{S_P-1}\right)\left[\dfrac{1}{1-\gamma S_P}-\dfrac{1}{1-\gamma}\right]
		\end{align}
		where $J\in\{R, C\}$, and the terms $M_R, M_C, S_R, S_C, S_P, \zeta_0$ are given in Lemma \ref{theorem_1}. Additionally, if $\boldsymbol{\pi}_\infty^*$ is a solution of (\ref{problem_2a}) with $\zeta = -G_C^0$, then the following inequalities hold $\forall N\geq N_0$ whenever $\gamma S_P<1$.
		\begin{align}
			\label{eq_17_thm2}
			&V_N^{*}(\boldsymbol{x}_0^N, 0) - V_N^{R}(\boldsymbol{x}_0^N, \boldsymbol{\pi}^*_{\infty}) \leq  2G_R^0 + G_C^0\left[\dfrac{4}{\zeta_0}\left(\dfrac{M_R}{1-\gamma}\right)\right],\\
			\label{eq_18_thm2}
			\text{and }& V_N^{C}(\boldsymbol{x}_0^N, \boldsymbol{\pi}_{\infty}^*) \leq 0
		\end{align}
		where $V_N^{R}(\cdot, \cdot)$, and $V_N^{C}(\cdot, \cdot)$ are defined in $(\ref{eq_def_v_r})$, and $(\ref{eq_def_v_c})$ respectively.
	\end{theorem}
	
	Inequality $(\ref{eq_threorem_2_1})$ states that the optimal value function obtained by solving (\ref{problem_1}) is approximated by the optimal value function obtained by solving (\ref{problem_2a}) with $\zeta=-G_C$ within an error of $\mathcal{O}(\sqrt{|\mathcal{X}|}/\sqrt{N})$. On the other hand, $(\ref{eq_17_thm2}), (\ref{eq_18_thm2})$ suggest that if the optimal policy sequence obtained by solving (\ref{problem_2a}), with $\zeta=-G_C$ is adopted in an $N$-agent system, then the value generated in such a system is at most $\mathcal{O}(\sqrt{|\mathcal{X}|}/\sqrt{N})$ distance away from the optimal $N$-agent value function. Moreover, the said policy does not violate the constraint of (\ref{problem_1}).

	Note that, the dependence of the approximation error on $N$ is still $\mathcal{O}(1/\sqrt{N})$. However, its dependence on the sizes of state, and action spaces has been reduced to $\mathcal{O}(\sqrt{|\mathcal{X}|})$ from $\mathcal{O}(\sqrt{|\mathcal{X}|}+\sqrt{|\mathcal{U}|})$ stated in Lemma \ref{theorem_1}. Therefore, the stated approximation result may be useful in those situations where reward, cost, and transition functions are independent of the action distribution, and the size of action space of individual agents is large. Interestingly, we could not derive an approximation error that is independent of the size of the state space, $|\mathcal{X}|$, by imposing the restriction that $r, c$, and $P$ are independent of the state distribution. This indicates an inherent asymmetry between the roles played by state, and action spaces in mean-field approximation.

	\section{Natural Policy Gradient Algorithm to Solve CMFC Problem}
	\label{sec:algo}
	
	In section \ref{sec:approximation_cmarl_via_cmfc}, we demonstrated that the $N$-agent problem (\ref{problem_1}) is well-approximated by the mean-field problem (\ref{problem_2a}) with appropriate choice of $\zeta$. In this section, we shall discuss how one can employ a Natural Policy Gradient (NPG) based algorithm to approximately solve (\ref{problem_2a}). Recall that in a mean-field setup, it is sufficient to track only one representative agent. At time $t$, the representative chooses an action $u_t\in \mathcal{U}$ based upon its observation of its own state, $x_t\in \mathcal{X}$, and the mean-field state distribution, $\boldsymbol{\mu}_t^{\infty}\in\mathcal{P}(\mathcal{X})$. Thus, (\ref{problem_2a}) can be described as a constrained single agent problem with state space $\mathcal{X}\times \mathcal{P}(\mathcal{X})$, and action space, $\mathcal{U}$. Without loss of generality, we can therefore assume the policy-sequences to be stationary \citep{dolgov2005stationary}. With slight abuse of notations, we denote both an arbitrary policy, and its associated stationary sequence by the same notation, $\pi$. The class of all admissible policies is, $\Pi$. Let, the elements of $\Pi$ be parameterized by the parameter, $\Phi\in \mathbb{R}^{\mathrm{d}}$. For a given policy, $\pi_{\Phi}$, the $Q$-function, $Q_{\Phi}^R(\cdot, \cdot, \cdot)$, value function, $V_{\Phi}^R(\cdot, \cdot)$, and the advantage function, $A_{\Phi}^R(\cdot, \cdot, \cdot)$ are defined as follows.
	\begin{align}
		\label{eq:def_q_phi}
		&Q_{\Phi}^R(x, \boldsymbol{\mu}, u)\triangleq \mathbb{E}\left[\sum_{t=0}^\infty\gamma^tr(x_t, u_t, \boldsymbol{\mu}_t^\infty, \boldsymbol{\nu}_t^{\infty})\Big| x_0 = x, \boldsymbol{\mu}_0=\boldsymbol{\mu}, u_0=u\right],\\
		\label{eq:def_v_phi}
		& V_{\Phi}^R(x, \boldsymbol{\mu})\triangleq \mathbb{E}\left[Q^R_{\Phi}(x, \boldsymbol{\mu}, u)\right],~~
		A_{\Phi}^R(x, \boldsymbol{\mu}, u) = Q_{\Phi}(x, \boldsymbol{\mu}, u) -  V_{\Phi}^R(x, \boldsymbol{\mu})
	\end{align}
	
	The expectation in $(\ref{eq:def_q_phi})$ is obtained over $x_{t+1}\sim P(x_t, u_t, \boldsymbol{\mu}_t, \boldsymbol{\nu}_t)$, $u_t\sim \pi_{\Phi}(x_t,\boldsymbol{\mu}_t)$, $\forall t\in \{0,1, \cdots\}$. The deterministic quantities, $\{\boldsymbol{\mu}_t^\infty, \boldsymbol{\nu}_t^\infty\}_{t\in \{0,1,\cdots\}}$ are evaluated using relations $(\ref{eq:mu_t_1})$, and $(\ref{eq:nu_t})$ respectively. On the other hand, the expectation in $(\ref{eq:def_v_phi})$ is computed over, $u\sim \pi_{\Phi}(x, \boldsymbol{\mu})$. The functions, $Q_{\Phi}^C$, $V_{\Phi}^C$, and $A_{\Phi}^C$ are defined similarly for the cost function, $c$. In the parametric form, (\ref{problem_2a}) can be rewritten as the following primal problem.
	\begin{align}
		\tag{PRIMAL}
		\label{problem_primal}
		\begin{split}
			\sup_{\Phi\in \mathbb{R}^{\mathrm{d}}}&~V_{\infty}^{R}(\boldsymbol{\mu}_0, \pi_{\Phi})~~~~
			\mathrm{subject~to}: ~V_{\infty}^{C}(\boldsymbol{\mu}_0, \pi_{\Phi}) \leq \zeta
		\end{split}
	\end{align}
	
	Its corresponding dual problem is as follows.
	\begin{align}
		\tag{DUAL}
		\label{problem_dual}
		\sup_{\Phi\in \mathbb{R}^{\mathrm{d}}} \inf_{\lambda\geq 0} V_{\infty}^{R}(\boldsymbol{\mu}_0, \pi_{\Phi}) + \lambda\left[\zeta - V_{\infty}^{C}(\boldsymbol{\mu}_0, \pi_{\Phi})\right]
	\end{align}
	
	Let $\Phi^*\in\mathbb{R}^{\mathrm{d}}$ be a solution of (\ref{problem_dual}). Ideally, $\Phi^*$ should be obtained via the following primal-dual natural policy gradient (NPG) updates that start from $\Phi_0$, $\lambda_0=0$ and utilize $\eta_1, \eta_2>0$ as learning parameters \citep{ding2020natural}.
	\begin{align}
		\label{npg_update}
		&\Phi_{j+1} = \Phi_j + \dfrac{\eta_1}{1-\gamma} \mathbf{w}_j,  \mathbf{w}_j\triangleq{\arg\min}_{\mathbf{w}\in\mathbb{R}^{\mathrm{d}}} ~E_{ \xi_{\boldsymbol{\mu}_0}^{\Phi_j}}(\mathbf{w},\Phi_j, \lambda_j),\\
		\label{lambda_update}
		&\lambda_{j+1} = \mathrm{P}_{[0, \infty)}\left[\lambda_j + \eta_2 [V_{\infty}^C(\boldsymbol{\mu}_0, \pi_{\Phi_j})-\zeta]\right]
	\end{align}
	
	The function, $\mathrm{P}_{[0, \infty)}$ projects its argument onto the set $[0, \infty)$. The term  $\xi_{\boldsymbol{\mu}_0}^{\Phi_j}$ indicates the occupancy measure induced by policy, $\pi_{\Phi_j}$ from the initial distribution, $\boldsymbol{\mu}_0$.
	The function, $E_{ \xi_{\boldsymbol{\mu}_0}^{\Phi_j}}(\cdot, \cdot, \cdot)$ and the occupancy measure $\xi_{\boldsymbol{\mu}_0}^{\Phi_j}$ are mathematically defined below.\vspace{-0.2cm}
		\begin{align}
			&\xi_{\boldsymbol{\mu}_0}^{\Phi_j} \triangleq \sum_{\tau=0}^{\infty}\gamma^{\tau} (x, \boldsymbol{\mu}, u) \mathbb{P}(x_\tau=x,\boldsymbol{\mu}_{\tau}=\boldsymbol{\mu},u_\tau=u
			|\boldsymbol{\mu}_0=\boldsymbol{\mu},\pi_{\Phi_j})(1-\gamma),\\
			&E_{ \xi_{\boldsymbol{\mu}_0}^{\Phi_j}}(\mathbf{w},\Phi_j, \lambda_j)\triangleq \mathbb{E}_{(x,\boldsymbol{\mu},u)\sim \xi_{\boldsymbol{\mu}_0}^{\Phi_j}}\Big[\Big(A_{\Phi_j}^{\lambda_j}(x,\boldsymbol{\mu},u)-\mathbf{w}^{\mathrm{T}}\nabla_{\Phi_j}\log \pi_{\Phi_j}(x,\boldsymbol{\mu})(u) \Big)^2\Big]
		\end{align}
		where $A_{\Phi_j}^{\lambda_j}(\cdot, \cdot, \cdot)$ is defined as $A_{\Phi_j}^{\lambda_j}\triangleq A_{\Phi_j}^R - \lambda_j A_{\Phi_j}^C$. Due to the sampling-based nature of the algorithm, we focus on the following approximate updates.
    \begin{align}
		\label{npg_update_approx}
		&\Phi_{j+1} = \Phi_j + \dfrac{\eta_1}{1-\gamma} \hat{\mathbf{w}}_j,\\
		\label{lambda_update_approx}
		&\lambda_{j+1} = \mathrm{P}_{[0, \infty)}\left[\lambda_j + \eta_2 [\hat{V}_{\infty}^C(\boldsymbol{\mu}_0, \pi_{\Phi_j})-\zeta]\right]
	\end{align}
    where $\hat{V}_{\infty}^C(\boldsymbol{\mu}_0, \pi_{\Phi_j})$ is an unbiased estimate of $V_{\infty}^C(\boldsymbol{\mu}_0, \pi_{\Phi_j})$, which is obtained via Algorithm \ref{algo_2} (Appendix \ref{sec:appndx_sampling}). To obtain $\hat{\mathbf{w}_j}$ (approximation of $\mathbf{w}_j$), we need to solve another minimization problem. We apply the stochastic gradient descent (SGD) updates described as follows to solve this sub-problem: $\mathbf{w}_{j,l+1}=\mathbf{w}_{j,l}-\alpha\mathbf{h}_{j,l}$ where the gradient $\mathbf{h}_{j,l}$ is given below \citep{ding2020natural}, and $\alpha>0$ is the learning parameter.
	\begin{align}
		\mathbf{h}_{j,l}\triangleq \Bigg(\mathbf{w}_{j,l}^{\mathrm{T}}\nabla_{\Phi_j}\log \pi_{\Phi_j}(x,\boldsymbol{\mu})(u)-\hat{A}^{\lambda_j}_{\Phi_j}(x,\boldsymbol{\mu},u)\Bigg)
		\nabla_{\Phi_j}\log \pi_{\Phi_j}(x,\boldsymbol{\mu})(u)
		\label{sub_prob_grad_update}
	\end{align}
	where $(x, \boldsymbol{\mu}, u)\sim \xi_{\boldsymbol{\mu}_0}^{\Phi_j}$, and $\hat{A}_{\Phi_j}^{\lambda_j}$ is an unbiased estimator of $A_{\Phi_j}^{\lambda_j}\triangleq A_{\Phi_j}^R-\lambda_j A_{\Phi_j}^C$. The sampling process is detailed in Algorithm \ref{algo_2} (Appendix \ref{sec:appndx_sampling}). Algorithm \ref{algo_1} summarizes the NPG process.

 	\begin{algorithm}
		\caption{Natural Policy Gradient Algorithm to solve the Dual Problem}
		\label{algo_1}
		\textbf{Input:} $\eta_1,\eta_2,\alpha$: Learning rates, $J,L$: Number of execution steps\\
		\hspace{1.3cm}$\mathbf{w}_0,\Phi_0, \lambda_0=0$: Initial parameters, $\boldsymbol{\mu}_0$: Initial state distribution, Constraint bound: $\zeta$
		
		\begin{algorithmic}[1]
			\FOR{$j\in\{0,1,\cdots,J-1\}$}
			{
				\STATE $\mathbf{w}_{j,0}\gets \mathbf{w}_0$\\
				\FOR {$l\in\{0,1,\cdots,L-1\}$}
				{
					\STATE Sample $(x,\boldsymbol{\mu},u)\sim\zeta_{\boldsymbol{\mu}_0}^{\Phi_j}$ and $\hat{A}_{\Phi_j}^{\lambda_j}(x,\boldsymbol{\mu},u)$ using Algorithm \ref{algo_2} \\
					\STATE Compute $\mathbf{h}_{j,l}$ using $(\ref{sub_prob_grad_update})$\\
					$\mathbf{w}_{j,l+1}\gets\mathbf{w}_{j,l}-\alpha\mathbf{h}_{j,l}$
				}
				\ENDFOR
				\STATE	$\hat{\mathbf{w}}_j\gets\frac{1}{L}\sum_{l=1}^{L}\mathbf{w}_{j,l}$\\
				\STATE	$\Phi_{j+1}\gets \Phi_j +\dfrac{\eta_1}{1-\gamma} \hat{\mathbf{w}}_j$\\
				\STATE Sample $(x,\boldsymbol{\mu},u)\sim\zeta_{\boldsymbol{\mu}_0}^{\Phi_j}$ and $\hat{V}_{\Phi_j}^{C}(x,\boldsymbol{\mu})$ using Algorithm \ref{algo_2} \\
				\STATE $\lambda_{j+1} \gets \mathrm{P}_{[0, \infty)}\left[\lambda_j + \eta_2 [\hat{V}_{\Phi_j}^C(x, \boldsymbol{\mu})-\zeta]\right]$
			}
			\ENDFOR
		\end{algorithmic}
		\textbf{Output:} $\{\Phi_1,\cdots,\Phi_J\}$: Policy parameters
	\end{algorithm}

 Recall that Lemma \ref{theorem_1} proves that the optimal solution of (\ref{problem_2a}) with appropriate choice of $\zeta$ approximately solves the $N$-agent problem (\ref{problem_1}). However, Algorithm \ref{algo_1} can only approximately solve (\ref{problem_2a}). One, therefore, naturally asks whether the solution of Algorithm \ref{algo_1} is close to the optimal solution of (\ref{problem_1}). Theorem \ref{npg_theorem} (stated below) provides an affirmative answer to this question. The proof of Theorem \ref{npg_theorem} is relegated to Appendix \ref{sec:appndx_npg_theorem}. The following assumptions are needed to establish this result.

	\begin{assumption}
		\label{ass_6}
		There exists $\zeta_1<0$ such that the (\ref{problem_2a}) problem associated with $\zeta=\zeta_1$ has a feasible solution.
	\end{assumption}
	
	\begin{assumption}
		\label{ass_7}
		$\forall \Phi\in\mathbb{R}^{\mathrm{d}}$, $\forall \boldsymbol{\mu}\in\mathcal{P}(\mathcal{X})$, $\forall x\in\mathcal{X}$, $\forall u\in\mathcal{U}$, 
		\begin{align*}
			\left|\nabla_{\Phi}\log\pi_{\Phi}(x,\boldsymbol{\mu})(u)\right|_1\leq G
		\end{align*}
		for some positive constant $G$.
	\end{assumption}
	
	\begin{assumption}
		\label{ass_8}
		$\forall \Phi_1,\Phi_2\in\mathbb{R}^{\mathrm{d}}$, $\forall \boldsymbol{\mu}\in\mathcal{P}(\mathcal{X})$,  $\forall x\in\mathcal{X}$, $\forall u\in\mathcal{U}$,
		\begin{align*}
			|\nabla_{\Phi_1}\log\pi_{\Phi_1}(x,\boldsymbol{\mu})(u)-\nabla_{\Phi_2}\log\pi_{\Phi_2}(x,\boldsymbol{\mu})(u)|_1
			\leq M|\Phi_1-\Phi_2|_1
		\end{align*}
		for some positive constant $M$.
	\end{assumption}
	
	\begin{assumption}
		\label{ass_9}
		$\forall \Phi\in\mathbb{R}^{\mathrm{d}}$, $\forall \boldsymbol{\mu}_0\in\mathcal{P}(\mathcal{X})$, 
		\begin{align*}
			E_{\xi_{\boldsymbol{\mu}_0}^{\Phi^*}}(\mathbf{w}^{*}_{\Phi},\Phi)\leq \epsilon_{\mathrm{bias}}, ~~\mathbf{w}^*_{\Phi}\triangleq{\arg\min}_{\mathbf{w}\in\mathbb{R}^{\mathrm{d}}}E_{\xi_{\boldsymbol{\mu}_0}^{\Phi}}(\mathbf{w},\Phi) 
		\end{align*}
		where $\Phi^*$ is the parameter of the optimal policy.
	\end{assumption}
	
	\begin{assumption}
		\label{ass_10}
		The gradient iterates $\{\mathbf{w}_{j,l}\}_{j\in\{0,\cdots, J-1\}, l\in\{0,\cdots, L-1\}}$, and $\{\mathbf{w}_j\}_{j\in \{0,\cdots, J-1\}}$ of Algorithm \ref{algo_1} are such that,
		\begin{align*}
			\mathbb{E}\left[|\mathbf{w}_{j,l}|_1^2\right]\leq W_0^2, \text{and}~\mathbb{E}\left[|\mathbf{w}_{j}|_1^2\right]\leq W_1^2
		\end{align*}
		for some constants $W_0$, $W_1$.
	\end{assumption}
	 Assumption \ref{ass_6} expresses Slater's condition for the (\ref{problem_2a}) problem. Assumption \ref{ass_7}, and \ref{ass_8} ensure that the log-likelihood function is Lipschitz and smooth with respect to the parameters. Assumption \ref{ass_9} says that the expressivity error of the parameterized policy class is bounded by a term $\epsilon_{\mathrm{bias}}$. If the policy class is complete (e.g., in softmax parameterization), then $\epsilon_{\mathrm{bias}}=0$. Although, in general, $\epsilon_{\mathrm{bias}}>0$, for rich policy classes (e.g., where policies are represented by dense/wide neural networks), its value is negligibly small. Finally, Assumption \ref{ass_10} dictates that the gradient estimates used in our algorithm are bounded. All of the above assumptions are standard in the policy gradient literature. For a detailed discussion on their validity, see \cite{ding2020natural}. 
	\begin{theorem}
		\label{npg_theorem}
		Let $\boldsymbol{x}_0^N$ be the initial joint state in an $N$-agent system, and $\boldsymbol{\mu}_0$ its empirical distribution. Let $\{\Phi_j\}_{j=1}^J$ denote the policy parameters yielded from Algorithm \ref{algo_1} for an initial parameter, $\Phi_0$, and $\zeta=-2G_C$. If Assumptions $\ref{assumption_1}-\ref{assumption_2}$, and $\ref{ass_6}-\ref{ass_10}$ hold, then for appropriate choices of $\eta_1, \eta_2, \alpha, J, L$, and sufficiently large $N$, the following relations hold when $\gamma S_P<1$.
		\begin{align}
			&\Big|~V_{N}^{*}(\boldsymbol{x}^N_0, 0)	-\dfrac{1}{J}\sum_{j=1}^{J}V_{\infty}^R(\boldsymbol{\mu}_0, \pi_{\Phi_j})\Big|\leq K_1 e + K_2\sqrt{\epsilon_{\mathrm{bias}}},\\
			& \dfrac{1}{J}\sum_{j=1}^JV_N^C(\boldsymbol{x}_0^N, \pi_{\Phi_j})\leq 0
		\end{align}
		where $\epsilon_{\mathrm{bias}}$ is defined in Assumption \ref{ass_9}, $K_1, K_2$ are constants, and $e\triangleq \frac{1}{\sqrt{N}}[\sqrt{|\mathcal{X}|}+\sqrt{|\mathcal{U}|}]$. The sample complexity of the process is $\mathcal{O}(e^{-6})$.
	\end{theorem}
	
	Theorem \ref{npg_theorem} states that the solution of (\ref{problem_2a}) with $\zeta=-2G_C$, obtained via Algorithm \ref{algo_1} closely  (within an error of $\mathcal{O}(e)$) approximates the optimal objective value obtained by solving the $N$-agent problem (\ref{problem_1}). Moreover, the obtained policy also satisfies the constraint of (\ref{problem_1}). The sample complexity of the whole process is $\mathcal{O}(e^{-6})$. Since the submission of this paper, some new developments have occurred in the field of algorithm design for CMDPs \citep{bai2023achieving,mondal2024sample}. In particular, \cite{mondal2024sample} have recently proposed an algorithm that can achieve $e+\sqrt{\epsilon_{\mathrm{bias}}}$ optimality error and zero constraint violation with $\Tilde{\mathcal{O}}(e^{-2})$ sample complexity. We believe that the result of Theorem \ref{npg_theorem} can be improved to $\Tilde{\mathcal{O}}(e^{-2})$ if one utilizes the algorithm suggested by \citep{mondal2024sample} for solving our CMARL problem. However, the verification of such a possibility is left for future research.
	
		\section{Experimental Results}
	\label{appndx_experiment}
	
	We consider the following setting (taken from \citep{subramanian2019reinforcement} with slight modifications) for our numerical experiment.  Consider a network of $N$ firms that produce the same product but with different qualities. At time $t\in\{0,1,\cdots\}$, the product quality of $i$-th, $i\in\{1,2,\cdots, N\}$, firm is denoted as $x_t^i$ which can assume value from the set $\mathcal{Q}\triangleq \{0, 1, \cdots, Q-1\}$. Each firm has two options. It can either remain unresponsive or invest some money to improve the quality of its product. These two possibilities are symbolised as the elements of the  action set, $\mathcal{U}\triangleq\{0,1\}$ where $0$ indicates unresponsiveness and $1$ denotes investment. If at time $t$, the $i$-th firm chooses the action, $u_t^i\in\mathcal{U}$, then its state in the next time step changes according to the following transition law.
	\begin{align*}
		x_{t+1}^i = 
		\begin{cases}
			x_t^i + \left\lfloor \chi\left(1-\dfrac{\boldsymbol{\bar{\mu}}_t^{N}}{Q-1}\right)  (Q-1-x_t^i)\right\rfloor~ \text{if}~ u_t^i = 1, \\
			x_t^i~~\hspace{4.4cm}\text{otherwise}
		\end{cases}
	\end{align*}
	where $\chi$ is a uniform random variable in $[0,1]$, and $\boldsymbol{\bar{\mu}}_t^{N}$ is empirical average product quality defined as the mean of the distribution, $\boldsymbol{\mu}_t^N$ defined in $(\ref{eq_4})$. Hence, if $u_t^i=0$ (unresponsiveness), then the product quality does not change. On the other hand, if the firm invests, i.e., $u_t^i=1$, then its product quality increases. The increase in the product quality, however, is dependent on the average product quality, $\boldsymbol{\bar{\mu}}_t^N$, in the economy. If $\boldsymbol{\bar{\mu}}_t^N$ is high, then it is difficult to improve the product quality of any individual firm. The reward, and cost received by the $i$-th firm at time $t$ are given as follows. 
	\begin{align*}
		&r(x_t^i, u_t^i,\boldsymbol{\mu}_t^{N}, \boldsymbol{\nu}_t^{N}) = \alpha_R x_t^i - \beta_R\boldsymbol{\bar{\mu}}_t^{N} -  \lambda_R u_t^i\\
		&c(x_t^i, u_t^i,\boldsymbol{\mu}_t^{N}, \boldsymbol{\nu}_t^{N}) = \lambda_C u_t^i
	\end{align*}

	\begin{figure*}[t]
		\begin{subfigure}{0.49\textwidth}
			\centering
			\includegraphics[width=\linewidth]{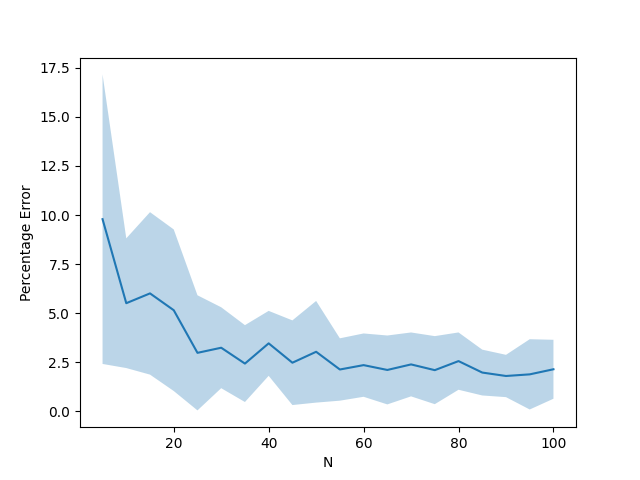}
			\caption{}
			\label{subfig_1a}
		\end{subfigure}
		\begin{subfigure}{0.49\textwidth}
			\centering
			\includegraphics[width=\linewidth]{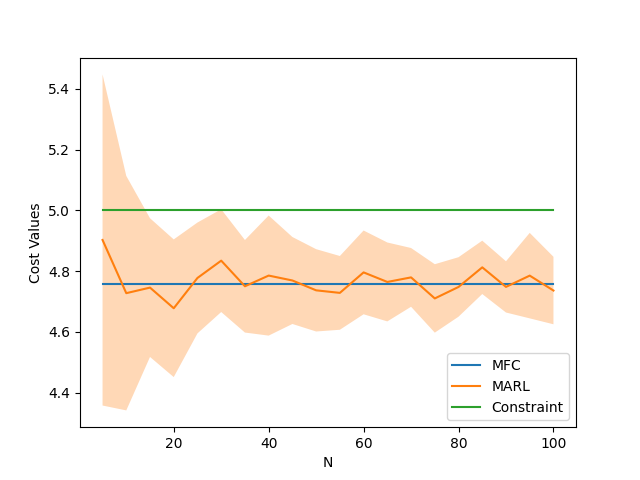}
			\caption{}
			\label{subfig_1b}
		\end{subfigure}
		\caption{Fig. (\ref{subfig_1a}) portrays the percentage error (defined by $(\ref{eq_error})$) as a function of $N$. On the other hand, Fig. (\ref{subfig_1b}) plots the $N$-agent (orange), and infinite-agent (blue) cost-values corresponding to the optimal mean-field policy as a function of $N$. It also shows that both of these values lie below the specified upper bound, $\zeta$ (green). The values of different  system parameters are given as: $\alpha_R=1$, $\beta_R=0.5$, $\lambda_R=0.5$, $\lambda_C=1$, $\zeta=5$, $\gamma=0.9$, and $Q=10$. The hyperparameters used in Algorithm \ref{algo_1} are chosen as follows: $\eta_1=\eta_2=\alpha=10^{-3}$, $J=L=10^2$.  The bold lines, and the half-width of the shaded regions respectively denote the mean values, and the standard deviation values obtained over $25$ random seeds. The experiments were performed on a $1.8$ GHz Dual-Core Intel $i5$ processor with $8$ GB $1600$ MHz DDR3 memory.}
		\label{fig_1}
	\end{figure*} 
	
	The reward consists of three parts. The first part, $\alpha_Rx_t^i$, is earned as  revenue; the second part, $\beta_R\boldsymbol{\bar{\mu}}_t^{N}$ expresses dependence on the whole population; the third part, $\lambda_R u_t^i$, is due to the investment. The cost is a constant, $\lambda_C$, if money is invested, otherwise it is zero. The objective of this $N$-agent RL problem is to maximize the expected time-discounted sum of rewards while ensuring that the cumulative time-discounted cost is bounded above by a certain constant, $\zeta$. Let $\pi^*_{\infty}$ be the optimal policy of its associated constraint mean-field control (CMFC) problem. In Fig. \ref{subfig_1a}, we demonstrate how the following error changes as a function of $N$.
	\begin{align}
		\label{eq_error}
		\mathrm{Error} = \left|\dfrac{V_N^R(\boldsymbol{x}_0^N, \pi^*_{\infty})-V_\infty^R(\boldsymbol{\mu}_0, \pi^*_{\infty})}{V_\infty^R(\boldsymbol{\mu}_0, \pi^*_{\infty})} \right|\times 100\%
	\end{align}
	The value functions $V_R^N$ and $V_\infty^R$ are defined in $(\ref{eq_def_v_r})$ and $(\ref{eq:v_r_mf})$, respectively. The initial state distribution, $\boldsymbol{\mu}_0$, is taken to be a uniform distribution over $\mathcal{Q}$, and $\boldsymbol{x}_0^N$ is obtained by taking $N$-independent samples from $\boldsymbol{\mu}_0$. The policy, $\pi^*_\infty$, is obtained via Algorithm \ref{algo_1}. Values of other relevant parameters are stated in Fig. \ref{fig_1}. We observe that the error decreases as a function of $N$. Essentially, Fig. \ref{subfig_1a} shows that if $N$ is large, then the $N$-agent cumulative average reward generated by $\pi^*_\infty$ is well approximated by the optimal mean-field value. In Fig. \ref{subfig_1b}, we exhibit that the $N$-agent and mean-field cost values generated by $\pi^*_{\infty}$ are close for large $N$, and both of them lie below the specified upper bound, $\zeta$.

	\section{Conclusions}
	\label{sec:conclusions}
	
	This paper shows that a constrained multi-agent reinforcement learning (CMARL) problem can be well-approximated via a constrained mean-field control (CMFC) problem with a suitably adjusted constraint bound. We have characterized the approximation error as a function of population size, and the sizes of state, and action spaces respectively. We also state an algorithm to solve the CMFC problem and analyze its sample complexity. One limitation of this study is that it only considers constraints in the form of time-discounted cumulative costs. Studying the same problem with other forms of constraints such as instantaneous cost, average cost, etc. could be a potential future direction.

    \newpage
	\appendix
	
	\section{Proof of Theorem \ref{corr_1}}
	\label{sec:appndx_corr_1}
	
	Using Lemma \ref{lemma_1a}, we obtain, 
	\begin{align}
        \label{eq_rand}
		V_N^C(\boldsymbol{x}_0^N, \boldsymbol{\pi}_\infty^*)\leq V_\infty^C(\boldsymbol{\mu}_0, \boldsymbol{\pi}_\infty^*)+G_C\overset{(a)}{\leq} 0
	\end{align}
	
	Inequality (a) follows from the fact that $\boldsymbol{\pi}^*_\infty$ is a feasible solution of (\ref{problem_2a}) with $\zeta=-G_C$. Due to $(\ref{eq_rand})$, $\boldsymbol{\pi}^*_\infty$ can be treated as a feasible solution of (\ref{problem_1}). Therefore,
	\begin{align*}
		&V_N^{R}(\boldsymbol{x}_0^N, \boldsymbol{\pi}_N^*) - V_N^{R}(\boldsymbol{x}_0^N, \boldsymbol{\pi}^*_{\infty}) \\
		&= V_N^{R}(\boldsymbol{x}^N_0, \boldsymbol{\pi}_N^*) - V_{\infty}^R(\boldsymbol{\mu}_0, \boldsymbol{\pi}_\infty^*) +
		V_{\infty}^R(\boldsymbol{\mu}_0, \boldsymbol{\pi}_\infty^*) -  V_N^{R}(\boldsymbol{x}_0^N, \boldsymbol{\pi}^*_{\infty}) \\
        &\overset{(a)}{=}V_N^*(\boldsymbol{x}_0^N, 0) - V_{\infty}^*(\boldsymbol{\mu}_0, -G_C)+
		V_{\infty}^R(\boldsymbol{\mu}_0, \boldsymbol{\pi}_\infty^*) -  V_N^{R}(\boldsymbol{x}_0^N, \boldsymbol{\pi}^*_{\infty})\\
		&\overset{(b)}{\leq} 2G_R + G_C\left[\dfrac{4}{\zeta_0}\left(\dfrac{M_R}{1-\gamma}\right)\right]
	\end{align*}
	
	Relation $(a)$ follows from the fact that $V_N^{R}(\boldsymbol{x}^N_0, \boldsymbol{\pi}_N^*)=V_N^*(\boldsymbol{x}_0^N, 0)$, and $V_{\infty}^R(\boldsymbol{\mu}_0, \boldsymbol{\pi}_\infty^*)=V_{\infty}^*(\boldsymbol{\mu}_0, -G_C)$. Please refer to Table \ref{table_notation} for a detailed explanation. Inequality (b) follows from Lemma \ref{theorem_1}, and Lemma \ref{lemma_1a}. This concludes the result.

	\section{Proof of Lemma \ref{lemma_1a}}
	\label{sec:appndx_lemma_1a}
	
	In order to establish Lemma \ref{lemma_1a}, the following results are necessary. We use the notation $\pi_t$ to indicate the $t$-th component of the policy sequence, $\boldsymbol{\pi}$. Moreover, the $\boldsymbol{\pi}$-induced empirical $N$-agent state, and action distributions at time $t$ are denoted as $\boldsymbol{\mu}_t^N, \boldsymbol{\nu}_t^N$ respectively. Their counterparts for infinite agent system are $\boldsymbol{\mu}_t^\infty$, and $\boldsymbol{\nu}_t^\infty$. Following the notation of section \ref{sec:cmarl}, the ($\boldsymbol{\pi}$-induced) joint state, and action at time $t$ are denoted as $\boldsymbol{x}_t^N\triangleq \{x_t^i\}_{i\in\{1,\cdots, N\}}, \boldsymbol{u}_t^N\triangleq\{u_t^i\}_{i\in\{1,\cdots,N\}}$ respectively.
	
	\subsection{Lipschitz Continuity Lemmas}
	\label{appndx_a1}

	\begin{lemma}
		\label{lemma_1}
		If $\nu^{\mathrm{MF}}(\cdot, \cdot)$ is defined by $(\ref{eq:nu_t})$, then the following holds $\forall t\in \{0,1,\cdots\}$.
		\begin{align}
			|\nu^{\mathrm{MF}}(\boldsymbol{\mu}_t^N, \pi_t)-\nu^{\mathrm{MF}}(\boldsymbol{\mu}_t^\infty, \pi_t)|_1\leq  (1+L_Q)|\boldsymbol{\mu}_t^N-\boldsymbol{\mu}_t^\infty|_1 
		\end{align}	
	\end{lemma}

	\begin{lemma}
		\label{lemma_2}
		If $P^{\mathrm{MF}}(\cdot, \cdot)$ is defined by $(\ref{eq:mu_t_1})$, then the following holds $\forall t\in \{0,1,\cdots\}$.
		\begin{align}
			|P^{\mathrm{MF}}(\boldsymbol{\mu}_t^N, \pi_t)-P^{\mathrm{MF}}(\boldsymbol{\mu}_t^\infty, \pi_t)|_1\leq S_P|\boldsymbol{\mu}-\boldsymbol{\bar{\mu}}|_1 
		\end{align}
		where $S_P\triangleq 1+2L_P+L_Q(1+L_P)$.
	\end{lemma}

	\begin{lemma}
		\label{lemma_3}
		If $r^{\mathrm{MF}}(\cdot, \cdot)$ is defined by $(\ref{eq:r_MF})$, then the following holds $\forall t\in \{0,1,\cdots\}$.
		\begin{align}
			|r^{\mathrm{MF}}(\boldsymbol{\mu}_t^N, \pi_t)-r^{\mathrm{MF}}(\boldsymbol{\mu}_t^\infty, \pi_t)|\leq S_R|\boldsymbol{\mu}_t^N-\boldsymbol{\mu}_t^\infty|_1 	\end{align}
		where $S_R\triangleq M_R + 2L_R +L_Q(M_R + L_R)$.
	\end{lemma}
	
	\begin{lemma}
		\label{lemma_3c}
		If $c^{\mathrm{MF}}(\cdot, \cdot)$ is defined by $(\ref{eq:c_MF})$, then the following holds $\forall t\in \{0,1,\cdots\}$.
		\begin{align}
			|c^{\mathrm{MF}}(\boldsymbol{\mu}_t^N, \pi_t)-c^{\mathrm{MF}}(\boldsymbol{\mu}_t^\infty, \pi_t)|\leq S_C|\boldsymbol{\mu}_t^N-\boldsymbol{\mu}_t^\infty|_1 	\end{align}
		where $S_C\triangleq M_C + 2L_C +L_Q(M_C + L_C)$.
	\end{lemma}
	
	Lemma $\ref{lemma_1}-\ref{lemma_3c}$ shows that the state and action evolution functions, average reward, and cost functions of an infinite agent system are Lipschitz continuous with respect to the state distribution argument. Lemma \ref{lemma_1} is an essential ingredient in the proof of Lemma \ref{lemma_2}, and \ref{lemma_3}. The proofs of Lemma \ref{lemma_1}, \ref{lemma_2}, and \ref{lemma_3} are presented in Appendix \ref{sec:appndx_lemma_1}, \ref{sec:appndx_lemma_2}, and \ref{sec:appndx_lemma_3} respectively. The proof of Lemma \ref{lemma_3c} is identical to that of Lemma \ref{lemma_3}.
	
	\subsection{Large-$N$ Approximation Lemmas}
	\label{appndx_a2}
	
	\begin{lemma}
		\label{lemma_4}
		The following inequality holds $\forall t\in \{0,1,\cdots\}$.
		\begin{align*}
			\mathbb{E}\big|\boldsymbol{\nu}^N_t - \nu^{\mathrm{MF}}(\boldsymbol{\mu}_t^N, \pi_t) \big|_1 \leq \dfrac{1}{\sqrt{N}}\sqrt{|\mathcal{U}|}
		\end{align*}
	\end{lemma}
	
	\begin{lemma}
		\label{lemma_5}
		The following inequality holds $\forall t\in \{0,1,\cdots\}$.
		\begin{align*}
			\mathbb{E}\left|\boldsymbol{\mu}^N_{t+1}-P^{\mathrm{MF}}(\boldsymbol{\mu}^N_t, \pi_t)\right|_1\leq \dfrac{C_P}{\sqrt{N}}\left[\sqrt{|\mathcal{X}|}+ \sqrt{|\mathcal{U}|}\right]
		\end{align*}
		where $C_P\triangleq 2+L_P$.
	\end{lemma}
	
	\begin{lemma}
		\label{lemma_6}
		The following inequality holds $\forall t\in \{0,1,\cdots\}$.
		\begin{align*}
			\mathbb{E}\left|\dfrac{1}{N}\sum_{i=1}r(x_t^i, u_t^i, \boldsymbol{\mu}_t^N, \boldsymbol{\nu}_t^N)-r^{\mathrm{MF}}(\boldsymbol{\mu}_t^N, \pi_t)\right|\leq \dfrac{M_R}{\sqrt{N}}+\dfrac{L_R}{\sqrt{N}}\sqrt{|\mathcal{U}|}
		\end{align*}
	\end{lemma}
	
	\begin{lemma}
		\label{lemma_6c}
		The following inequality holds $\forall t\in \{0,1,\cdots\}$.
		\begin{align*}
			\mathbb{E}\left|\dfrac{1}{N}\sum_{i=1}c(x_t^i, u_t^i, \boldsymbol{\mu}_t^N, \boldsymbol{\nu}_t^N)-c^{\mathrm{MF}}(\boldsymbol{\mu}_t^N, \pi_t)\right|\leq \dfrac{M_C}{\sqrt{N}}+\dfrac{L_C}{\sqrt{N}}\sqrt{|\mathcal{U}|}
		\end{align*}
	\end{lemma}
	
	The proofs of Lemma \ref{lemma_4}, \ref{lemma_5}, and \ref{lemma_6} are given in Appendix \ref{sec:appndx_lemma_4}, \ref{sec:appndx_lemma_5}, and \ref{sec:appndx_lemma_6}, respectively. The proof of Lemma \ref{lemma_6c} can be obtained by replacing $r,M_R,L_R$ with $c,M_C,L_C$ respectively in the proof of Lemma \ref{lemma_6}.  Finally, invoking Lemma \ref{lemma_2}, and \ref{lemma_5}, we obtain the following.
	
	\begin{lemma}
		\label{lemma_6a}
		The following inequality holds $\forall t\in \{0,1,\cdots\}$.
		\begin{align*}
			\mathbb{E}|\boldsymbol{\mu}_t^N-\boldsymbol{\mu}_t^{\infty}|\leq \dfrac{C_P}{\sqrt{N}}\left[\sqrt{|\mathcal{X}|}+\sqrt{|\mathcal{U}|}\right]\left(\dfrac{S_P^t-1}{S_P-1}\right)
		\end{align*}
		where $S_P$ is defined in Lemma \ref{lemma_2} while $C_P$ is given in Lemma \ref{lemma_5}.
	\end{lemma}
	
	The proof of Lemma \ref{lemma_6a} is given in Appendix \ref{appndx_6a}. 
	
	\subsection{Proof of the Lemma}
	\label{appndx:a3}
	
	We shall establish $(\ref{eq_19})$ only for $J=R$. The proof for $J=C$ is identical. Consider the following difference,
	\begin{align*}
		&|V_{N}^R(\boldsymbol{x}_0^N, \boldsymbol{\pi})-V_{\infty}^R(\boldsymbol{\mu}_0, \boldsymbol{\pi})|\\
		& \overset{(a)}{\leq} \sum_{t=0}^{\infty}\gamma^t\left|\dfrac{1}{N}\sum_{i=1}^N \mathbb{E}\left[r(x_t^i, u_t^i, \boldsymbol{\mu}_t^N, \boldsymbol{\nu}_t^N)\right]-r^{\mathrm{MF}}(\boldsymbol{\mu}_t^\infty, \pi_t)\right|\\
		&\leq \underbrace{\sum_{t=0}^{\infty}\gamma^t\mathbb{E}\left|\dfrac{1}{N}r(x_t^i, u_t^i, \boldsymbol{\mu}_t^N, \boldsymbol{\nu}_t^N)- r^{\mathrm{MF}}(\boldsymbol{\mu}_t^N, \pi_t)\right|}_{\triangleq J_1} + \underbrace{\sum_{t=0}^\infty \gamma^t\mathbb{E}\left|r^{\mathrm{MF}}(\boldsymbol{\mu}_t^N, \pi_t)-r^{\mathrm{MF}}(\boldsymbol{\mu}_t^\infty, \pi_t)\right|}_{\triangleq J_2}
	\end{align*}
	
	Inequality (a) follows from the definition of the functions $V_{N}^R(\cdot, \cdot)$, $V_{\infty}^R(\cdot, \cdot)$ given in $(\ref{eq_def_v_r})$, and $(\ref{eq:v_r_mf})$ respectively. The first term, $J_1$ can be bounded using Lemma \ref{lemma_6} as follows.
	\begin{align*}
		J_1\leq\left(\dfrac{1}{1-\gamma}\right)\left[\dfrac{M_R}{\sqrt{N}}+\dfrac{L_R}{\sqrt{N}}\sqrt{|\mathcal{U}|}\right]
	\end{align*}
	
	The second term, $J_2$, can be bounded as follows.
	\begin{align*}
		J_2&\triangleq\sum_{t=0}^{\infty}\gamma^t\mathbb{E}|r^{\mathrm{MF}}(\boldsymbol{\mu}_t^N, \pi_t)-r^{\mathrm{MF}}(\boldsymbol{\mu}_t^\infty, \pi_t)|\\
		&\overset{(a)}{\leq} S_R \sum_{t=0}^\infty \gamma^t \mathbb{E}|\boldsymbol{\mu}_t^N - \boldsymbol{\mu}_t^\infty|\\
		&\overset{(b)}{\leq} \dfrac{1}{\sqrt{N}}\left[\sqrt{|\mathcal{X}|}+\sqrt{|\mathcal{U}|}\right] \left(\dfrac{S_RC_P}{S_P-1}\right)\left[\dfrac{1}{1-\gamma S_P}-\dfrac{1}{1-\gamma}\right]
	\end{align*}
	Inequality (a) follows from Lemma \ref{lemma_3}, whereas (b) is a consequence of Lemma \ref{lemma_6a}.

	\section{Proof of Theorem \ref{theorem_2}}
	\label{appndx_theorem_2}
	
	The proof of Theorem \ref{theorem_2} hinges on the following Lemma.
	\begin{lemma}
		\label{lemma_2a}
		Let $\boldsymbol{x}_0^N$ be the initial joint state in an $N$-agent system, and $\boldsymbol{\mu}_0$ its empirical distribution. If Assumption $\ref{assumption_1}-\ref{assumption_2}$ hold, then there exists $N_0>0$ such that $\forall N\geq N_0$, and $\forall \boldsymbol{\pi}\in\Pi_{\infty}$, the following inequalities hold whenever $\gamma S_P<1$.
		\begin{align}
			\label{eq_19a}
			&|V_N^J(\boldsymbol{x}_0^N, \boldsymbol{\pi}) - V_{\infty}^J(\boldsymbol{\mu}_0, \boldsymbol{\pi})|\leq G_J^0
		\end{align}
		where $J\in \{R, C\}$. The value functions $V_N^{R}, V_N^{C}$ are given by $(\ref{eq_def_v_r}), (\ref{eq_def_v_c})$ respectively, and the terms $G_{R}^0, G_{C}^0, S_P$ are defined in Theorem \ref{theorem_2}.
	\end{lemma}
	
	The role of Lemma \ref{lemma_2a} in establishing Theorem \ref{theorem_2} is analogous to the role of Lemma \ref{lemma_1a} in proving Lemma \ref{theorem_1}. In this section, we shall primarily focus on proving Lemma \ref{lemma_2a}. Once it is established, Theorem \ref{theorem_2} can be proven following an argument similar to that is used in section \ref{sec:proof_theorem_1}.  
	
	\subsection{Auxiliary Lemmas}
	
	To prove Lemma \ref{lemma_2a}, the following results are necessary. The notations are the same as used in Appendix \ref{sec:appndx_lemma_1a}.
	
	\begin{lemma}
		\label{lemma_8}
		The following inequality holds $\forall t\in \{0,1,\cdots\}$.
		\begin{align}
			\mathbb{E}\left|\boldsymbol{\mu}^N_{t+1}-P^{\mathrm{MF}}(\boldsymbol{\mu}^N_t, \pi_t)\right|_1\leq \dfrac{2}{\sqrt{N}} \sqrt{|\mathcal{X}|}
		\end{align}
	\end{lemma}
	
	\begin{lemma}
		\label{lemma_9}
		The following inequality holds $\forall t\in \{0,1,\cdots\}$.
		\begin{align*}
			\mathbb{E}\left|\dfrac{1}{N}\sum_{i=1}r(x_t^i, u_t^i, \boldsymbol{\mu}_t^N)-r^{\mathrm{MF}}(\boldsymbol{\mu}_t^N, \pi_t)\right|\leq \dfrac{M_R}{\sqrt{N}}
		\end{align*}
	\end{lemma}
	
	\begin{lemma}
		\label{lemma_9a}
		The following inequality holds $\forall t\in \{0,1,\cdots\}$.
		\begin{align*}
			\mathbb{E}\left|\dfrac{1}{N}\sum_{i=1}c(x_t^i, u_t^i, \boldsymbol{\mu}_t^N)-c^{\mathrm{MF}}(\boldsymbol{\mu}_t^N, \pi_t)\right|\leq \dfrac{M_C}{\sqrt{N}}
		\end{align*}
	\end{lemma}
	
	\begin{lemma}
		\label{lemma_10}
		The following inequality holds $\forall t\in \{0,1,\cdots\}$.
		\begin{align*}
			\mathbb{E}|\boldsymbol{\mu}_t^N-\boldsymbol{\mu}_t^{\infty}|\leq \dfrac{2}{\sqrt{N}}\sqrt{|\mathcal{X}|}\left(\dfrac{S_P^t-1}{S_P-1}\right)
		\end{align*}
		The term $S_P$ is defined in Lemma \ref{lemma_2}.
	\end{lemma} 
	
	The proofs of Lemma \ref{lemma_8}, \ref{lemma_9},  \ref{lemma_10} are relegated to Appendix $\ref{appndx_lemma_8}$, $\ref{appndx_lemma_9}$, and $\ref{appndx_lemma_10}$, respectively. The proof of Lemma \ref{lemma_9a} can be obtained by replacing $r$ with $c$, and $M_R$ with $M_C$ in the proof of Lemma \ref{lemma_9}. 
	
	\subsection{Proof of Lemma \ref{lemma_2a}}
	\label{appndx_b2}
	
	We use the same notations as in Appendix \ref{appndx:a3}. Consider the following difference,
	\begin{align*}
		&|V_{N}^R(\boldsymbol{x}_0^N, \boldsymbol{\pi})-V_{\infty}^R(\boldsymbol{\mu}_0, \boldsymbol{\pi})|\\
		& \overset{(a)}{\leq} \sum_{t=0}^{\infty}\gamma^t\left|\dfrac{1}{N}\sum_{i=1}^N r(x_t^i, u_t^i, \boldsymbol{\mu}_t^N)\right]-\mathbb{E}\left[r^{\mathrm{MF}}(\boldsymbol{\mu}_t^\infty, \pi_t)\right|\\
		&\leq \underbrace{\sum_{t=0}^{\infty}\gamma^t\mathbb{E}\left|\dfrac{1}{N}r(x_t^i, u_t^i, \boldsymbol{\mu}_t^N)- r^{\mathrm{MF}}(\boldsymbol{\mu}_t^N, \pi_t)\right|}_{\triangleq J_1} + \underbrace{\sum_{t=0}^\infty \gamma^t\mathbb{E}\left|r^{\mathrm{MF}}(\boldsymbol{\mu}_t^N, \pi_t)-r^{\mathrm{MF}}(\boldsymbol{\mu}_t^\infty, \pi_t)\right|}_{\triangleq J_2} 
	\end{align*}
	
	Inequality (a) follows from the definition of the value functions $V_{N}^R(\cdot, \cdot)$, $V_{\infty}^C(\cdot, \cdot)$ given in $(\ref{eq_def_v_r})$, and $(\ref{eq:v_r_mf})$ respectively. The first term, $J_1$ can be bounded using Lemma \ref{lemma_6} as follows.
	\begin{align*}
		J_1\leq \dfrac{M_R}{\sqrt{N}}\left(\dfrac{1}{1-\gamma}\right)
	\end{align*}
	
	The second term, $J_2$, can be bounded as follows.
	\begin{align*}
		J_2&\triangleq\sum_{t=0}^{\infty}\gamma^t\mathbb{E}|r^{\mathrm{MF}}(\boldsymbol{\mu}_t^N, \pi_t)-r^{\mathrm{MF}}(\boldsymbol{\mu}_t^\infty, \pi_t)|\\
		&\overset{(a)}{\leq} S_R \sum_{t=0}^\infty \gamma^t \mathbb{E}|\boldsymbol{\mu}_t^N - \boldsymbol{\mu}_t^\infty|
		\overset{(b)}{\leq} \dfrac{1}{\sqrt{N}}\sqrt{|\mathcal{X}|} \left(\dfrac{2S_R}{S_P-1}\right)\left[\dfrac{1}{1-\gamma S_P}-\dfrac{1}{1-\gamma}\right]
	\end{align*}
	Inequality (a) follows from Lemma \ref{lemma_3}, whereas (b) is a consequence of Lemma \ref{lemma_10}. This establishes $(\ref{eq_19a})$ for $J=R$. The other case, $J=C$, can be proven similarly.

	\section{Proof of Lemma \ref{lemma_1}}
	\label{sec:appndx_lemma_1}
	
	Note the following inequalities.
	\begin{align*}
		&|\nu^{\mathrm{MF}}(\boldsymbol{\mu}_t^N, \pi_t)-\nu^{\mathrm{MF}}(\boldsymbol{\mu}_t^\infty, \pi_t)|_1\\
		&\overset{(a)}{=}\Bigg|\sum_{x\in \mathcal{X}}\pi_t(x, \boldsymbol{\mu}_t^N)\boldsymbol{\mu}_t^N(x) - \sum_{x\in \mathcal{X}} \pi_t(x, \boldsymbol{\mu}_t^\infty)\boldsymbol{\mu}_t^\infty(x)\Bigg|_1\\
		&=\sum_{u\in \mathcal{U}}\Bigg|\sum_{x\in \mathcal{X}}\pi_t(x, \boldsymbol{\mu}_t^N)(u)\boldsymbol{\mu}_t^N(x) - \sum_{x\in \mathcal{X}} \pi_t(x, \boldsymbol{\mu}_t^\infty)(u)\boldsymbol{\mu}_t^\infty(x)\Bigg|\\
		&\leq \sum_{x\in \mathcal{X}}\sum_{u\in \mathcal{U}}|\pi_t(x, \boldsymbol{\mu}_t^N)(u)\boldsymbol{\mu}_t^N(x) -  \pi_t(x, \boldsymbol{\mu}_t^\infty)(u)\boldsymbol{\mu}_t^\infty(x)|\\
		&\leq \sum_{x\in \mathcal{X}}|\boldsymbol{\mu}_t^N(x)-\boldsymbol{\mu}_t^\infty(x)|\underbrace{\sum_{u\in \mathcal{U}}\pi_t(x, \boldsymbol{\mu}_t^N)(u)}_{=1} + \sum_{x\in \mathcal{X}}\boldsymbol{\mu}_t^\infty(x)\sum_{u\in \mathcal{U}}|\pi_t(x, \boldsymbol{\mu}_t^N)(u)-\pi_t(x, \boldsymbol{\mu}_t^\infty)(u)|\\
		&\overset{(b)}{\leq}|\boldsymbol{\mu}_t^N-\boldsymbol{\mu}_t^\infty|_1 + L_Q|\boldsymbol{\mu}_t^N-\boldsymbol{\mu}_t^\infty|\underbrace{\sum_{x\in \mathcal{X}}\boldsymbol{\mu}_t^\infty(x)}_{=1} 
		\overset{(c)}{=} (1+L_Q)|\boldsymbol{\mu}_t^N-\boldsymbol{\mu}_t^\infty|_1 
	\end{align*}
	
	Equality (a) follows from the definition of $\nu^{\mathrm{MF}}(\cdot, \cdot)$ as given in $(\ref{eq:nu_t})$. On the other hand, inequality (b) is a consequence of Assumption \ref{assumption_policy}, and the fact that $\pi_t(x, \boldsymbol{\mu}_t^N)$ is probability distribution $\forall x\in \mathcal{X}$. Finally, (c) follows because $\boldsymbol{\mu}_t^\infty$ is a probability distribution. This concludes the result.
	
	\section{Proof of Lemma \ref{lemma_2}}
	\label{sec:appndx_lemma_2}
	
	Note that, 
	\begin{align*}
		|P^{\mathrm{MF}}&(\boldsymbol{\mu}_t^N, \pi_t)-P^{\mathrm{MF}}(\boldsymbol{\mu}_t^\infty,\pi_t)|_1\\
		&\overset{(a)}{=}\Bigg| \sum_{x\in \mathcal{X}}\sum_{u\in \mathcal{U}}P(x, u, \boldsymbol{\mu}_t^N, \nu^{\mathrm{MF}}(\boldsymbol{\mu}_t^N,\pi_t))\pi_t(x, \boldsymbol{\mu}_t^N)(u)\boldsymbol{\mu}_t^N(x)\\
        &\hspace{2cm}-  P(x, u, \boldsymbol{\mu}_t^\infty, \nu^{\mathrm{MF}}(\boldsymbol{\mu}_t^\infty,\pi_t))\pi_t(x, \boldsymbol{\mu}_t^\infty)(u)\boldsymbol{\mu}_t^\infty(x)\Bigg|_1
		\leq J_1 + J_2
	\end{align*}
	
	Equality (a) follows from the definition of $P^{\mathrm{MF}}(\cdot,\cdot)$ as stated in $(\ref{eq:mu_t_1})$. The first term can be bounded as follows.
	\begin{align*}
		J_1\triangleq& \sum_{x\in \mathcal{X}}\sum_{u\in \mathcal{U}}\Big|P(x, u, \boldsymbol{\mu}_t^N, \nu^{\mathrm{MF}}(\boldsymbol{\mu}_t^N, \pi_t))-P(x, u, \boldsymbol{\mu}_t^\infty, \nu^{\mathrm{MF}}(\boldsymbol{\mu}_t^\infty, \pi_t))\Big|_1\times\pi_t(x, \boldsymbol{\mu}_t^N)(u)\boldsymbol{\mu}_t^N(x)\\
		&\overset{(a)}{\leq}L_P\left[|\boldsymbol{\mu}_t^N-\boldsymbol{\mu}_t^\infty|_1+|\nu^{\mathrm{MF}}(\boldsymbol{\mu}_t^N, \pi_t)-\nu^{\mathrm{MF}}(\boldsymbol{\mu}_t^\infty, \pi_t)|_1\right] \times \underbrace{\sum_{x\in \mathcal{X}}\boldsymbol{\mu}_t^N(x)\sum_{u\in \mathcal{U}}\pi_t(x, \boldsymbol{\mu}_t^N)(u)}_{=1}\\
		&\overset{(b)}{\leq} L_P(2+L_Q)|\boldsymbol{\mu}_t^N-\boldsymbol{\mu}_t^\infty|_1
	\end{align*}
	
	Inequality $(a)$ is a result of Assumption \ref{assumption_1}(e) while $(b)$ follows from Lemma \ref{lemma_1}, and the fact that $\pi_t(x, \boldsymbol{\mu}_t^N)$, $\boldsymbol{\mu}_t^N$ are probability distributions $\forall x\in \mathcal{X}$. The second term, $J_2$ can be bounded as follows.
	\begin{align*}
		J_2\triangleq &\sum_{x\in \mathcal{X}} \sum_{u\in \mathcal{U}} \underbrace{|P(x, u, \boldsymbol{\mu}_t^\infty, \nu^{\mathrm{MF}}(\boldsymbol{\mu}_t^\infty, \pi_t))|_1}_{=1} \times |\pi_t(x, \boldsymbol{\mu}_t^N)(u)\boldsymbol{\mu}_t^N(x) - \pi_t(x, \boldsymbol{\mu}_t^\infty)(u)\boldsymbol{\mu}_t^\infty(x)|\\
		&\leq \sum_{x\in \mathcal{X}}|\boldsymbol{\mu}_t^N(x)-\boldsymbol{\mu}_t^\infty(x)|\underbrace{\sum_{u\in \mathcal{U}}\pi_t(x, \boldsymbol{\mu}_t^N)(u)}_{=1} \\
        &\hspace{1cm}+ \sum_{x\in \mathcal{X}}\boldsymbol{\mu}_t^\infty(x)\sum_{u\in \mathcal{U}}|\pi_t(x, \boldsymbol{\mu}_t^N)(u)-\pi_t(x, \boldsymbol{\mu}_t^\infty)(u)|\\
		&\overset{(a)}{\leq}|\boldsymbol{\mu}_t^N-\boldsymbol{\mu}_t^\infty|_1 + L_P|\boldsymbol{\mu}_t^N-\boldsymbol{\mu}_t^\infty|_1\underbrace{\sum_{x\in \mathcal{X}}\boldsymbol{\mu}_t^\infty(x)}_{=1}  \\
        &= (1+L_P)|\boldsymbol{\mu}_t^N-\boldsymbol{\mu}_t^\infty|_1
	\end{align*}
	
	Inequality $(a)$ follows from the fact that $\pi_t(x, \boldsymbol{\mu}_t^N)$ is a probability distribution $\forall x\in \mathcal{X}$ while $(b)$ uses Assumption \ref{assumption_policy}, and the fact that $\boldsymbol{\mu}_t^\infty$ is a probability distribution. This concludes the result.
	
	\section{Proof of Lemma \ref{lemma_3}}
	\label{sec:appndx_lemma_3}
	Notice the following inequality, 
	\begin{align*}
		&|r^{\mathrm{MF}}(\boldsymbol{\mu}_t^N, \pi_t)-r^{\mathrm{MF}}(\boldsymbol{\mu}_t^\infty,\pi_t)|\\
		&\overset{(a)}{=}\Bigg| \sum_{x\in \mathcal{X}}\sum_{u\in \mathcal{U}}r(x, u, \boldsymbol{\mu}_t^N, \nu^{\mathrm{MF}}(\boldsymbol{\mu}_t^N,\pi_t))\pi_t(x, \boldsymbol{\mu}_t^N)(u)\boldsymbol{\mu}_t^N(x)\\
        &\hspace{2cm}-  r(x, u, \boldsymbol{\mu}_t^\infty, \nu^{\mathrm{MF}}(\boldsymbol{\mu}_t^\infty,\pi_t))\pi_t(x, \boldsymbol{\mu}_t^\infty)(u)\boldsymbol{\mu}_t^\infty(x)\Bigg|\leq J_1 + J_2
	\end{align*}
	
	Equality (a) follows from the definition of $r^{\mathrm{MF}}(\cdot,\cdot)$ as depicted in $(\ref{eq:r_MF})$. The first term can be bounded as follows.
	\begin{align*}
		J_1\triangleq& \sum_{x\in \mathcal{X}}\sum_{u\in \mathcal{U}}\Big|r(x, u, \boldsymbol{\mu}_t^N, \nu^{\mathrm{MF}}(\boldsymbol{\mu}_t^N, \pi_t))-r(x, u, \boldsymbol{\mu}_t^\infty, \nu^{\mathrm{MF}}(\boldsymbol{\mu}_t^\infty, \pi_t))\Big|\times\pi_t(x, \boldsymbol{\mu}_t^N)(u)\boldsymbol{\mu}_t^N(x)\\
		&\overset{(a)}{\leq}L_R\left[|\boldsymbol{\mu}_t^N-\boldsymbol{\mu}_t^\infty|_1+|\nu^{\mathrm{MF}}(\boldsymbol{\mu}_t^N, \pi_t)-\nu^{\mathrm{MF}}(\boldsymbol{\mu}_t^\infty, \pi_t)|_1\right] \times \underbrace{\sum_{x\in \mathcal{X}}\boldsymbol{\mu}_t^N(x)\sum_{u\in \mathcal{U}}\pi_t(x, \boldsymbol{\mu}_t^N)(u)}_{=1}\\
		&\overset{(b)}{\leq} L_R(2+L_Q)|\boldsymbol{\mu}_t^N-\boldsymbol{\mu}_t^\infty|_1 
	\end{align*}
	
	Inequality $(a)$ is a result of Assumption \ref{assumption_1}(c) while relation $(b)$ follows from Lemma \ref{lemma_1}, and the fact that $\pi_t(x, \boldsymbol{\mu}_t^\infty)$, $\boldsymbol{\mu}_t^N$ are probability distributions $\forall x\in \mathcal{X}$. The term, $J_2$ can be bounded as follows.
	\begin{align*}
		J_2\triangleq &\sum_{x\in \mathcal{X}} \sum_{u\in \mathcal{U}} |r(x, u, \boldsymbol{\mu}_t^\infty, \nu^{\mathrm{MF}}(\boldsymbol{\mu}_t^\infty, \pi_t))| \times |\pi_t(x, \boldsymbol{\mu}_t^N)(u)\boldsymbol{\mu}_t^N(x) - \pi_t(x, \boldsymbol{\mu}_t^\infty)(u)\boldsymbol{\mu}_t^\infty(x)|\\
		&\overset{(a)}{\leq} M_R \sum_{x\in \mathcal{X}}\sum_{u\in \mathcal{U}} |\pi_t(x, \boldsymbol{\mu}_t^N)(u)\boldsymbol{\mu}_t^N(x) - \pi_t(x, \boldsymbol{\mu}_t^\infty)(u)\boldsymbol{\mu}_t^\infty(x)|\\
		&\leq  M_R\sum_{x\in \mathcal{X}}|\boldsymbol{\mu}_t^N(x)-\boldsymbol{\mu}_t^\infty(x)|\underbrace{\sum_{u\in \mathcal{U}}\pi_t(x, \boldsymbol{\mu}_t^N)(u)}_{=1} \\
        &\hspace{1cm}+ M_R\sum_{x\in \mathcal{X}}\boldsymbol{\mu}_t^\infty(x)\sum_{u\in \mathcal{U}}|\pi_t(x, \boldsymbol{\mu}_t^N)(u)-\pi_t(x, \boldsymbol{\mu}_t^\infty)(u)|\\
		&\overset{(b)}{\leq} M_R|\boldsymbol{\mu}_t^N-\boldsymbol{\mu}_t^\infty|_1 +M_RL_Q|\boldsymbol{\mu}_t^N-\boldsymbol{\mu}_t^\infty|_1\underbrace{\sum_{x\in\mathcal{X}}\boldsymbol{\mu}_t^\infty(x)}_{=1}  \overset{(c)}{=} M_R(1+L_Q)|\boldsymbol{\mu}_t^N-\boldsymbol{\mu}_t^\infty|_1 
	\end{align*}
	
	Inequality $(a)$ follows from Assumption \ref{assumption_1}(a). On the other hand, $(b)$ is a consequence of Assumption \ref{assumption_policy}. Finally, (c) follows from the fact that $\boldsymbol{\mu}_t^\infty$ is a probability distribution. This concludes the result.
	
	\section{Proof of Lemma \ref{lemma_4}}
	\label{sec:appndx_lemma_4}
	
	The following Lemma is required to prove the result.
	
	\begin{lemma}
		\label{lemma_7}
		If $\forall m\in \{1, \cdots, M\}$, $\{X_{mn}\}_{n\in \{1,\cdots, N\}}$ are independent random variables that lie in $[0, 1]$, and satisfy $\sum_{m\in \{1, \cdots, M\}} \mathbb{E}[X_{mn}]\leq 1$, $\forall n\in \{1, \cdots, N\}$,  then the following holds,
		\begin{align}
			\sum_{m=1}^{M}\mathbb{E}\left|\sum_{n=1}^{N}\left(X_{mn}-\mathbb{E}[X_{mn}]\right)\right|\leq \sqrt{MN}
		\end{align}
	\end{lemma}
	
	Lemma \ref{lemma_7} is adapted from Lemma 13 of	\citep{mondal2021approximation}.
	
	Observe the following inequalities.
	\begin{align*}
		&\mathbb{E}\left|\boldsymbol{\nu}_t^N-\nu^{\mathrm{MF}}(\boldsymbol{\mu}_t^N, \pi_t)\right|_1\\
		&=\mathbb{E}\left[\mathbb{E}\left[\left|\boldsymbol{\nu}^N_t-\nu^{\mathrm{MF}}(\boldsymbol{\mu}^N_t, \pi_t)\right|_1\Big| \boldsymbol{x}_t^N\right]\right]\\
		&\overset{(a)}{=}\mathbb{E}\left[\mathbb{E}\left[\left|\boldsymbol{\nu}_t^N-\sum_{x\in \mathcal{X}}\pi_t(x, \boldsymbol{\mu}_t^N)\boldsymbol{\mu}_t^N(x)\right|_1\Bigg| \boldsymbol{x}_t^N\right]\right]\\
		&=\mathbb{E}\left[\mathbb{E}\left[\sum_{u\in \mathcal{U}}\left|\boldsymbol{\nu}^N_t(u)-\sum_{x\in \mathcal{X}}\pi_t(x, \boldsymbol{\mu}_t^N)(u)\boldsymbol{\mu}^N_t(x)\right|\Bigg| \boldsymbol{x}_t^N\right]\right]\\
		&\overset{(b)}{=}\mathbb{E}\left[\sum_{u\in \mathcal{U}}\mathbb{E}\left[\dfrac{1}{N}\left|\sum_{i=1}^N\delta(u_t^i=u)-\dfrac{1}{N}\sum_{x\in \mathcal{X}}\pi_t(x, \boldsymbol{\mu}_t^N)(u)\sum_{i=1}^N\delta(x_t^i=x)\right|\Bigg|\boldsymbol{x}_t^N\right]\right]\\
		&= \mathbb{E}\left[\sum_{u\in \mathcal{U}}\mathbb{E}\left[\left|\dfrac{1}{N}\sum_{i=1}^N\delta(u_t^i=u)-\dfrac{1}{N}\sum_{i=1}^N\pi_t(x_t^i, \boldsymbol{\mu}_t^N)(u)\right|\Bigg|\boldsymbol{x}_t^N\right]\right]\overset{(c)}{\leq} \dfrac{1}{\sqrt{N}}\sqrt{|\mathcal{U}|}
	\end{align*}
	
	Equality (a) follows from the definition of $\nu^{\mathrm{MF}}(\cdot, \cdot)$ as depicted in $(\ref{eq:nu_t})$ while (b) is a consequence of the definitions of $\boldsymbol{\mu}_t^N, \boldsymbol{\nu}_t^N$. Finally, (c) is an application of Lemma \ref{lemma_7}. Specifically, it uses the facts that, $\{u_t^i\}_{i\in \{1, \cdots, N\}}$ are conditionally independent given $\boldsymbol{x}_t^N$, and
	\begin{align*}
		&\mathbb{E}\left[\delta(u_t^i=u)\Big| \boldsymbol{x}_t^N\right] = \pi_t(x_t^i, \boldsymbol{\mu}_t^N)(u), ~~\sum_{u\in \mathcal{U}}\mathbb{E}\left[\delta(u_t^i=u)\Big| \boldsymbol{x}_t^N\right] = 1
	\end{align*}
	$~\forall i\in \{1, \cdots, N\},\forall u\in \mathcal{U}$. This concludes the lemma.
	
	\section{Proof of Lemma \ref{lemma_5}}
	\label{sec:appndx_lemma_5}
	
	Note that, 
	\begin{align*}
		&\mathbb{E}\left|\boldsymbol{\mu}_{t+1}^N-P^{\mathrm{MF}}(\boldsymbol{\mu}^N_t, \pi_t)\right|_1
		\overset{(a)}{=}\sum_{x\in \mathcal{X}}\mathbb{E}\bigg|\dfrac{1}{N}\sum_{i=1}^N\delta(x_{t+1}^i=x)\\
        &\hspace{1cm}-\sum_{x'\in \mathcal{X}}\sum_{u\in \mathcal{U}}P(x', u, \boldsymbol{\mu}^N_t, \nu^{\mathrm{MF}}(\boldsymbol{\mu}^N_{t}, \pi_t))(x)\pi_t(x', \boldsymbol{\mu}_t^N)(u)\dfrac{1}{N}\sum_{i=1}^N\delta(x_t^i=x')\bigg|\\
		&=\sum_{x\in \mathcal{X}}\mathbb{E}\left|\dfrac{1}{N}\sum_{i=1}^N\delta(x_{t+1}^i=x)-\dfrac{1}{N}\sum_{i=1}^N \sum_{u\in \mathcal{U}}P(x_t^i, u, \boldsymbol{\mu}^N_t, \nu^{\mathrm{MF}}(\boldsymbol{\mu}^N_t, \pi_t))(x)\pi_t(x_t^i, \boldsymbol{\mu}_t^N)(u)\right|\\
		&\leq J_1+J_2+J_3
	\end{align*}
	
	Equality (a) follows from the definition of $P^{\mathrm{MF}}(\cdot, \cdot)$ as given in $(\ref{eq:mu_t_1})$. The first term, $J_1$, can be upper bounded as follows.
	\begin{align*}
		J_1&\triangleq \dfrac{1}{N}\sum_{x\in \mathcal{X}}\mathbb{E}\left|\sum_{i=1}^N\delta(x_{t+1}^i=x)-\sum_{i=1}^N P(x_t^i, u_t^i, \boldsymbol{\mu}^N_t, \boldsymbol{\nu}_t^N)(x)\right|\\
		& = \dfrac{1}{N}\sum_{x\in \mathcal{X}}\mathbb{E}\left[\mathbb{E}\left[\left|\sum_{i=1}^N\delta(x_{t+1}^i=x)-\sum_{i=1}^N P(x_t^i, u_t^i, \boldsymbol{\mu}^N_t, \boldsymbol{\nu}_t^N)(x)\right|\Bigg|\boldsymbol{x}_t^N, \boldsymbol{u}_t^N\right]\right]\overset{(a)}{\leq} \dfrac{1}{\sqrt{N}}\sqrt{|\mathcal{X}|}
	\end{align*}
	
	Inequality $(a)$ can be derived using Lemma \ref{lemma_7}, and the facts that $\{x_{t+1}^i\}_{i\in \{1, \cdots, N\}}$ are conditionally independent given $\{\boldsymbol{x}_t^N, \boldsymbol{u}_t^N\}$, and,
	\begin{align*}
		&\mathbb{E}\left[\delta(x_{t+1}^i=x)\Big| \boldsymbol{x}_t^N, \boldsymbol{u}_t^N\right] = P(x_t^i, u_t^i, \boldsymbol{\mu}^N_t, \boldsymbol{\nu}_t^N)(x),~~\sum_{x\in \mathcal{X}} \mathbb{E}\left[\delta(x_{t+1}^i=x)\Big| \boldsymbol{x}_t^N, \boldsymbol{u}_t^N\right] = 1
	\end{align*}
	$\forall i\in \{1, \cdots, N\}$, and $\forall x\in \mathcal{X}$. The second term can be upper bounded as follows.
	\begin{align*}
		J_2&\triangleq \dfrac{1}{N}\sum_{x\in \mathcal{X}}\mathbb{E}\left|\sum_{i=1}^NP(x_t^i, u_t^i, \boldsymbol{\mu}_t^N, \boldsymbol{\nu}_t^N)(x)- \sum_{i=1}^N P(x_t^i, u_t^i, \boldsymbol{\mu}^N_t, \nu^{\mathrm{MF}}(\boldsymbol{\mu}^N_t, \pi_t))(x)\right|\\
		&\leq \dfrac{1}{N}\sum_{i=1}^N \mathbb{E}\left|P(x_t^i, u_t^i, \boldsymbol{\mu}_t^N, \boldsymbol{\nu}_t^N)- P(x_t^i, u_t^i, \boldsymbol{\mu}^N_t, \nu^{\mathrm{MF}}(\boldsymbol{\mu}^N_t, \pi_t))\right|_1\\
		&\overset{(a)}{\leq} L_P\mathbb{E}\left|\boldsymbol{\nu}_t^N-\nu^{\mathrm{MF}}(\boldsymbol{\mu}_t^N, \pi_t)\right|_1
		\overset{(b)}{\leq} \dfrac{L_P}{\sqrt{N}}\sqrt{|\mathcal{U}|}
	\end{align*}
	
	Inequality (a) follows from Assumption \ref{assumption_2} while (b) follows from Lemma \ref{lemma_4}. Finally, the third term can be bounded as follows.
	\begin{align*}
		J_3&\triangleq \dfrac{1}{N}\sum_{x\in \mathcal{X}}\mathbb{E}\bigg|\sum_{i=1}^N P(x_t^i, u_t^i, \boldsymbol{\mu}^N_t, \nu^{\mathrm{MF}}(\boldsymbol{\mu}^N_t, \pi_t))(x) \\
        &\hspace{1cm}- \sum_{i=1}^N\sum_{u\in \mathcal{U}} P(x_t^i, u, \boldsymbol{\mu}^N_t, \nu^{\mathrm{MF}}(\boldsymbol{\mu}^N_t, \pi_t))(x)\pi_t(x_t^i, \boldsymbol{\mu}_t^N)(u) \bigg|\overset{(a)}{\leq} \dfrac{1}{\sqrt{N}}\sqrt{|\mathcal{X}|}
	\end{align*} 
	
	Inequality (a) is a consequence of Lemma \ref{lemma_7}. Specifically, it uses the facts that, $\{u_t^i\}_{i\in \{1, \cdots, N\}}$ are conditionally independent given $\boldsymbol{x}_t^N$, and
	\begin{align*}
		&\mathbb{E}\left[P(x_t^i, u_t^i, \boldsymbol{\mu}^N_t, \nu^{\mathrm{MF}}(\boldsymbol{\mu}^N_t, \pi_t))(x) \Big| \boldsymbol{x}_t^N\right] = \sum_{u\in \mathcal{U}} P(x_t^i, u, \boldsymbol{\mu}^N_t, \nu^{\mathrm{MF}}(\boldsymbol{\mu}^N_t, \pi_t))(x)\pi_t(x_t^i, \boldsymbol{\mu}_t^N)(u),	\\
		& \sum_{x\in \mathcal{X}} \mathbb{E}\left[P(x_t^i, u_t^i, \boldsymbol{\mu}^N_t, \nu^{\mathrm{MF}}(\boldsymbol{\mu}^N_t, \pi_t))(x) \Big| \boldsymbol{x}_t^N\right] = 1
	\end{align*}
	$\forall i\in \{1,\cdots, N\}$, and $\forall x\in\mathcal{X}$. This concludes the Lemma.
	
	\section{Proof of Lemma \ref{lemma_6}}
	\label{sec:appndx_lemma_6}
	
	Note that,  
	\begin{align*}
		&\mathbb{E}\left|\dfrac{1}{N}\sum_{i=1}r(x_t^i, u_t^i, \boldsymbol{\mu}_t^N, \boldsymbol{\nu}_t^N)-r^{\mathrm{MF}}(\boldsymbol{\mu}_t^N, \pi_t)\right|\\
		&\overset{(a)}{=}\mathbb{E}\left|\dfrac{1}{N}\sum_{i=1}^Nr(x_t^i, u_t^i, \boldsymbol{\mu}_t^N, \boldsymbol{\nu}_t^N)-\sum_{x\in \mathcal{X}}\sum_{u\in \mathcal{U}}r(x, u, \boldsymbol{\mu}_t^N, \nu^{\mathrm{MF}}(\boldsymbol{\mu}_t^N, \pi_t))\pi_t(x, \boldsymbol{\mu}_t^N)(u)\dfrac{1}{N}\sum_{i=1}^N\delta(x_t^i=x)\right|\\
		&=\mathbb{E}\left|\dfrac{1}{N}\sum_{i=1}^Nr(x_t^i, u_t^i, \boldsymbol{\mu}_t^N, \boldsymbol{\nu}_t^N)-\dfrac{1}{N}\sum_{i=1}^N\sum_{u\in \mathcal{U}}r(x_t^i, u, \boldsymbol{\mu}_t^N, \nu^{\mathrm{MF}}(\boldsymbol{\mu}_t^N, \pi_t))\pi_t(x_t^i,  \boldsymbol{\mu}_t^N)(u)\right|\\
		&\leq J_1 + J_2
	\end{align*}
	
	Equation (a) is a result of the definition of $r^{\mathrm{MF}}(\cdot, \cdot)$ as given in $(\ref{eq:r_MF})$. The first term, $J_1$, can be bounded as follows.
	\begin{align*}
		J_1&\triangleq \dfrac{1}{N}\mathbb{E}\left|\sum_{i=1}^N r(x_t^i, u_t^i, \boldsymbol{\mu}_t^N, \boldsymbol{\nu}_t^N)-\sum_{i=1}^N r(x_t^i, u_t^i, \boldsymbol{\mu}_t^N, \nu^{\mathrm{MF}}(\boldsymbol{\mu}_t^N, \pi_t))\right|\\
		&\leq  \dfrac{1}{N}\mathbb{E}\sum_{i=1}^N \left|r(x_t^i, u_t^i, \boldsymbol{\mu}_t^N, \boldsymbol{\nu}_t^N)- r(x_t^i, u_t^i, \boldsymbol{\mu}_t^N, \nu^{\mathrm{MF}}(\boldsymbol{\mu}_t^N, \pi_t))\right|\\
		&\overset{(a)}{\leq} L_R\mathbb{E}\left|\boldsymbol{\nu}_t^N-\nu^{\mathrm{MF}}(\boldsymbol{\mu}_t^N, \pi_t)\right|_1\overset{(b)}{\leq} \dfrac{L_R}{\sqrt{N}}\sqrt{|\mathcal{U}|}
	\end{align*}
	
	Inequality (a) follows from Assumption \ref{assumption_1}, and (b) is a consequence of Lemma \ref{lemma_4}. The second term, $J_2$, can be bounded as follows.
	\begin{align*}
		&J_2\triangleq \dfrac{1}{N} \mathbb{E} \left|\sum_{i=1}^N r(x_t^i, u_t^i, \boldsymbol{\mu}_t^N, \nu^{\mathrm{MF}}(\boldsymbol{\mu}_t^N, \pi_t)) - \sum_{i=1}^N\sum_{u\in \mathcal{U}} r(x_t^i, u, \boldsymbol{\mu}_t^N, \nu^{\mathrm{MF}}(\boldsymbol{\mu}_t^N, \pi_t))\pi_t(x_t^i, \boldsymbol{\mu}_t^N)(u)\right|\\
		& =\dfrac{1}{N} \mathbb{E}\left[\mathbb{E}\left[ \left|\sum_{i=1}^N r(x_t^i, u_t^i, \boldsymbol{\mu}_t^N, \nu^{\mathrm{MF}}(\boldsymbol{\mu}_t^N, \pi_t)) - \sum_{i=1}^N\sum_{u\in \mathcal{U}} r(x_t^i, u, \boldsymbol{\mu}_t^N, \nu^{\mathrm{MF}}(\boldsymbol{\mu}_t^N, \pi_t))\pi_t(x_t^i, \boldsymbol{\mu}_t^N)(u)\right|\Big|\boldsymbol{x}_t^N\right]\right]\\
		& =\dfrac{M_R}{N} \mathbb{E}\left[\mathbb{E}\left[ \left|\sum_{i=1}^N r_0(x_t^i, u_t^i, \boldsymbol{\mu}_t^N, \nu^{\mathrm{MF}}(\boldsymbol{\mu}_t^N, \pi_t)) - \sum_{i=1}^N\sum_{u\in \mathcal{U}} r_0(x_t^i, u, \boldsymbol{\mu}_t^N, \nu^{\mathrm{MF}}(\boldsymbol{\mu}_t^N, \pi_t))\pi_t(x_t^i, \boldsymbol{\mu}_t^N)(u)\right|\Big|\boldsymbol{x}_t^N\right]\right]\\
		&\overset{(a)}{\leq}\dfrac{M_R}{\sqrt{N}}
	\end{align*}
	
	where $r_0(\cdot, \cdot, \cdot, \cdot)\triangleq r(\cdot, \cdot, \cdot, \cdot)/M_R$. Inequality (a) follows from Lemma \ref{lemma_7}. Specifically, it uses the fact that $\{u_t^i\}_{i\in \{1, \cdots, N\}}$ are conditionally independent given $\boldsymbol{x}_t$, and
	\begin{align*}
		&|r_0(x_t^i, u_t^i, \boldsymbol{\mu}_t^N, \nu^{\mathrm{MF}}(\boldsymbol{\mu}_t^N, \pi_t))|\leq 1,\\	
		&\mathbb{E}\left[r_0(x_t^i, u_t^i, \boldsymbol{\mu}_t^N, \nu^{\mathrm{MF}}(\boldsymbol{\mu}_t^N, \pi_t))\Big| \boldsymbol{x}_t^N\right] = \sum_{u\in \mathcal{U}} r_0(x_t^i, u, \boldsymbol{\mu}_t^N, \nu^{\mathrm{MF}}(\boldsymbol{\mu}_t^N, \pi_t))\pi_t(x_t^i, \boldsymbol{\mu}_t^N)(u)
	\end{align*}
	$\forall i\in \{1, \cdots, N\}, \forall u\in \mathcal{U}$.
	
	\section{Proof of Lemma \ref{lemma_6a}}
	\label{appndx_6a}
	
	Observe that, 
	\begin{align*}
		\mathbb{E}|\boldsymbol{\mu}_t^N-\boldsymbol{\mu}_t^\infty|_1
		&\leq \mathbb{E}\left|\boldsymbol{\mu}_t^N-P^{\mathrm{MF}}(\boldsymbol{\mu}_{t-1}^N, \pi_{t-1})\right|_1 +  \mathbb{E}\left|P^{\mathrm{MF}}(\boldsymbol{\mu}_{t-1}^N, \pi_{t-1}) - \boldsymbol{\mu}_t^\infty\right|_1\\
		&\overset{(a)}{\leq} \dfrac{C_P}{\sqrt{N}}\left[\sqrt{|\mathcal{X}|}+\sqrt{|\mathcal{U}|}\right]+ \mathbb{E}\left|P^{\mathrm{MF}}(\boldsymbol{\mu}_{t-1}^N, \pi_{t-1}) - P^{\mathrm{MF}}(\boldsymbol{\mu}^\infty_{t-1}, \pi_{t-1})\right|_1
	\end{align*}
	
	Inequality (a) follows from Lemma \ref{lemma_5}, and relation $(\ref{eq:mu_t_1})$. Using Lemma \ref{lemma_2}, we get
	\begin{align*}
		&\left|P^{\mathrm{MF}}(\boldsymbol{\mu}_{t-1}^N, \pi_{t-1}) - P^{\mathrm{MF}}(\boldsymbol{\mu}^\infty_{t-1}, \pi_{t-1})\right|_1
		\overset{(a)}{\leq}S_P|\boldsymbol{\mu}_{t-1}^N-\boldsymbol{\mu}_{t-1}^\infty|_1
	\end{align*}
	Combining, we get,
	\begin{align}
		\mathbb{E}|\boldsymbol{\mu}_t^N-\boldsymbol{\mu}_t^\infty|_1\leq \dfrac{C_P}{\sqrt{N}}\left[\sqrt{|\mathcal{X}|}+\sqrt{|\mathcal{U}|}\right]+ S_P\mathbb{E}\left|\boldsymbol{\mu}_{t-1}^N-\boldsymbol{\mu}^\infty_{t-1}\right|_1
	\end{align}
	
	Recursively applying the above inequality, we finally obtain,
	\begin{align*}
		\mathbb{E}|\boldsymbol{\mu}_{t}^N-\boldsymbol{\mu}^\infty_{t}|_1\leq \dfrac{C_P}{\sqrt{N}}\left[\sqrt{|\mathcal{X}|}+\sqrt{|\mathcal{U}|}\right]\left(\dfrac{S_P^t-1}{S_P-1}\right)
	\end{align*}

	\section{Proof of Lemma \ref{lemma_8}}
	\label{appndx_lemma_8}
	
	Note that, 
	\begin{align*}
		&\mathbb{E}\left|\boldsymbol{\mu}_{t+1}^N-P^{\mathrm{MF}}(\boldsymbol{\mu}^N_t, \pi_t)\right|_1\\
		&\overset{(a)}{=}\sum_{x\in \mathcal{X}}\mathbb{E}\left|\dfrac{1}{N}\sum_{i=1}^N\delta(x_{t+1}^i=x)-\sum_{x'\in \mathcal{X}}\sum_{u\in \mathcal{U}}P(x', u, \boldsymbol{\mu}^N_t)(x)\pi_t(x', \boldsymbol{\mu}_t^N)(u)\dfrac{1}{N}\sum_{i=1}^N\delta(x_t^i=x')\right|\\
		&=\sum_{x\in \mathcal{X}}\mathbb{E}\left|\dfrac{1}{N}\sum_{i=1}^N\delta(x_{t+1}^i=x)-\dfrac{1}{N}\sum_{i=1}^N \sum_{u\in \mathcal{U}}P(x_t^i, u, \boldsymbol{\mu}^N_t)(x)\pi_t(x_t^i, \boldsymbol{\mu}_t^N)(u)\right|\leq J_1+J_2
	\end{align*}
	
	Equality (a) follows from the definition of $P^{\mathrm{MF}}(\cdot, \cdot)$ as depicted in $(\ref{eq:mu_t_1})$. The first term, $J_1$, can be upper bounded as follows.
	\begin{align*}
		J_1&\triangleq \dfrac{1}{N}\sum_{x\in \mathcal{X}}\mathbb{E}\left|\sum_{i=1}^N\delta(x_{t+1}^i=x)-\sum_{i=1}^N P(x_t^i, u_t^i, \boldsymbol{\mu}^N_t)(x)\right|\\
		& = \dfrac{1}{N}\sum_{x\in \mathcal{X}}\mathbb{E}\left[\mathbb{E}\left[\left|\sum_{i=1}^N\delta(x_{t+1}^i=x)-\sum_{i=1}^N P(x_t^i, u_t^i, \boldsymbol{\mu}^N_t)(x)\right|\Bigg|\boldsymbol{x}_t^N, \boldsymbol{u}_t^N\right]\right]\overset{(a)}{\leq} \dfrac{1}{\sqrt{N}}\sqrt{|\mathcal{X}|}
	\end{align*}
	
	Inequality $(a)$ can be derived using Lemma \ref{lemma_7}, and the facts that $\{x_{t+1}^i\}_{i\in \{1, \cdots, N\}}$ are conditionally independent given $\{\boldsymbol{x}^N_t, \boldsymbol{u}^N_t\}$, and,
	\begin{align*}
		&\mathbb{E}\left[\delta(x_{t+1}^i=x)\Big| \boldsymbol{x}^N_t, \boldsymbol{u}^N_t\right] = P(x_t^i, u_t^i, \boldsymbol{\mu}^N_t)(x),~~\sum_{x\in \mathcal{X}} \mathbb{E}\left[\delta(x_{t+1}^i=x)\Big| \boldsymbol{x}_t^N, \boldsymbol{u}_t^N\right] = 1
	\end{align*}
	$\forall i\in \{1, \cdots, N\}$, and $\forall x\in \mathcal{X}$. The second term can be bounded as follows.
	\begin{align*}
		J_2&\triangleq \dfrac{1}{N}\sum_{x\in \mathcal{X}}\mathbb{E}\left|\sum_{i=1}^N P(x_t^i, u_t^i, \boldsymbol{\mu}^N_t)(x) - \sum_{i=1}^N\sum_{u\in \mathcal{U}} P(x_t^i, u, \boldsymbol{\mu}^N_t)(x)\pi_t(x_t^i, \boldsymbol{\mu}_t^N)(u) \right|
		\overset{(a)}{\leq} \dfrac{\sqrt{|\mathcal{X}|}}{\sqrt{N}}
	\end{align*} 
	
	Inequality (a) is a consequence of Lemma \ref{lemma_7}. Specifically, it uses the facts that, $\{u_t^i\}_{i\in \{1, \cdots, N\}}$ are conditionally independent given $\boldsymbol{x}_t^N$, and
	\begin{align*}
		&\mathbb{E}\left[P(x_t^i, u_t^i, \boldsymbol{\mu}^N_t)(x) \Big| \boldsymbol{x}_t^N\right] = \sum_{u\in \mathcal{U}} P(x_t^i, u, \boldsymbol{\mu}^N_t)(x)\pi_t(x_t^i, \boldsymbol{\mu}_t^N)(u),	\\
		& \sum_{x\in \mathcal{X}} \mathbb{E}\left[P(x_t^i, u_t^i, \boldsymbol{\mu}^N_t)(x) \Big| \boldsymbol{x}_t^N\right] = 1
	\end{align*}
	$\forall i\in \{1,\cdots, N\}$, and $\forall x\in\mathcal{X}$. This concludes the Lemma.
	
	\section{Proof of Lemma \ref{lemma_9}}
	\label{appndx_lemma_9}
	
	Note that,  
	\begin{align*}
		&\mathbb{E}\left|\dfrac{1}{N}\sum_{i=1}r(x_t^i, u_t^i, \boldsymbol{\mu}_t^N)-r^{\mathrm{MF}}(\boldsymbol{\mu}_t^N, \pi_t)\right|\\
		&\overset{(a)}{=}\mathbb{E}\left|\dfrac{1}{N}\sum_{i=1}^Nr(x_t^i, u_t^i, \boldsymbol{\mu}_t^N)-\sum_{x\in \mathcal{X}}\sum_{u\in \mathcal{U}}r(x, u, \boldsymbol{\mu}_t^N)\pi_t(x, \boldsymbol{\mu}_t^N)(u)\dfrac{1}{N}\sum_{i=1}^N\delta(x_t^i=x)\right|\\
		&=\mathbb{E}\left|\dfrac{1}{N}\sum_{i=1}^Nr(x_t^i, u_t^i, \boldsymbol{\mu}_t^N)-\dfrac{1}{N}\sum_{i=1}^N\sum_{u\in \mathcal{U}}r(x_t^i, u, \boldsymbol{\mu}_t^N)\pi_t(x_t^i,  \boldsymbol{\mu}_t^N)(u)\right|\\
		& =\dfrac{1}{N} \mathbb{E}\left[\mathbb{E}\left[ \left|\sum_{i=1}^N r(x_t^i, u_t^i, \boldsymbol{\mu}_t^N) - \sum_{i=1}^N\sum_{u\in \mathcal{U}} r(x_t^i, u, \boldsymbol{\mu}_t^N)\pi_t(x_t^i, \boldsymbol{\mu}_t^N)(u)\right|\Big|\boldsymbol{x}_t^N\right]\right]\\
		& =\dfrac{M_R}{N} \mathbb{E}\left[\mathbb{E}\left[ \left|\sum_{i=1}^N r_0(x_t^i, u_t^i, \boldsymbol{\mu}_t^N) - \sum_{i=1}^N\sum_{u\in \mathcal{U}} r_0(x_t^i, u, \boldsymbol{\mu}_t^N)\pi_t(x_t^i, \boldsymbol{\mu}_t^N)(u)\right|\Big|\boldsymbol{x}_t^N\right]\right]
		\overset{(a)}{\leq}\dfrac{M_R}{\sqrt{N}}
	\end{align*}
	
	where $r_0(\cdot, \cdot, \cdot, \cdot)\triangleq r(\cdot, \cdot, \cdot, \cdot)/M_R$. Inequality (a) follows from Lemma \ref{lemma_7}. Specifically, it uses the fact that $\{u_t^i\}_{i\in \{1, \cdots, N\}}$ are conditionally independent given $\boldsymbol{x}_t^N$, and
	\begin{align*}
		|r_0(x_t^i, u_t^i, \boldsymbol{\mu}_t^N)|\leq 1, ~~\mathbb{E}\left[r_0(x_t^i, u_t^i, \boldsymbol{\mu}_t^N)\Big| \boldsymbol{x}_t^N\right] = \sum_{u\in \mathcal{U}} r_0(x_t^i, u, \boldsymbol{\mu}_t^N)\pi_t(x_t^i, \boldsymbol{\mu}_t^N)(u)
	\end{align*}
	$\forall i\in \{1, \cdots, N\}, \forall u\in \mathcal{U}$.
	
	\section{Proof of Lemma \ref{lemma_10}}
	\label{appndx_lemma_10}
	
	Observe that, 
	\begin{align*}
		\mathbb{E}|\boldsymbol{\mu}_t^N-\boldsymbol{\mu}_t^\infty|_1
		&\leq \mathbb{E}\left|\boldsymbol{\mu}_t^N-P^{\mathrm{MF}}(\boldsymbol{\mu}_{t-1}^N, \pi_{t-1})\right|_1 +  \mathbb{E}\left|P^{\mathrm{MF}}(\boldsymbol{\mu}_{t-1}^N, \pi_{t-1}) - \boldsymbol{\mu}_t^\infty\right|_1\\
		&\overset{(a)}{\leq} \dfrac{2}{\sqrt{N}}\sqrt{|\mathcal{X}|}+ \mathbb{E}\left|P^{\mathrm{MF}}(\boldsymbol{\mu}_{t-1}^N, \pi_{t-1}) - P^{\mathrm{MF}}(\boldsymbol{\mu}^\infty_{t-1}, \pi_{t-1})\right|_1
	\end{align*}
	
	Inequality (a) follows from Lemma \ref{lemma_5}, and relation $(\ref{eq:mu_t_1})$. Using Lemma \ref{lemma_2}, we get
	\begin{align*}
		&\left|P^{\mathrm{MF}}(\boldsymbol{\mu}_{t-1}^N, \pi_{t-1}) - P^{\mathrm{MF}}(\boldsymbol{\mu}^\infty_{t-1}, \pi_{t-1})\right|_1
		\overset{(a)}{\leq}S_P|\boldsymbol{\mu}_{t-1}^N-\boldsymbol{\mu}_{t-1}^\infty|_1
	\end{align*}
	Combining, we get,
	\begin{align}
		\mathbb{E}|\boldsymbol{\mu}_t^N-\boldsymbol{\mu}_t^\infty|_1\leq \dfrac{2}{\sqrt{N}}\sqrt{|\mathcal{X}|}+ S_P\mathbb{E}\left|\boldsymbol{\mu}_{t-1}^N-\boldsymbol{\mu}^\infty_{t-1}\right|_1 \leq \dfrac{2}{\sqrt{N}}\sqrt{|\mathcal{X}|}\left(\dfrac{S_P^t-1}{S_P-1}\right)
	\end{align}
	the last inequality is obtained via recursion.	
	
	\section{Proof of Theorem \ref{npg_theorem}}
	\label{sec:appndx_npg_theorem}
	
	Fix $J, L$. Following Theorem 3 of \citep{ding2020natural}, we can write,
	\begin{align}
		\label{eq_49}
		&\Big|V_{\infty}^{*}(\boldsymbol{\mu}_0, -2G_C)	-\dfrac{1}{J}\sum_{j=1}^{J}V_{\infty}^R(\boldsymbol{\mu}_0, \pi_{\Phi_j})\Big|\leq C_1\dfrac{1}{\sqrt{J}} + C_2\left(\sqrt{\epsilon_{\mathrm{bias}}}+\dfrac{C_3}{\sqrt{L}}\right)
	\end{align}
	
	for some constants $C_1, C_2, C_3$. On the other hand, Lemma \ref{theorem_1} suggests,
	\begin{align}
		\label{eq_15_apndx}
		|V_N^*(\boldsymbol{x}_0^N, 0) - V_{\infty}^*(\boldsymbol{\mu}_0, -G_C)| \leq G_R + G_C\left[\dfrac{4}{\zeta_0}\left(\dfrac{M_R}{1-\gamma}\right)\right]
	\end{align}
	
	Recall the definition of $\zeta_1$ from Assumption \ref{ass_6}. If $N$ is sufficiently large, then $2G_C<\zeta$. Using the concavity of $V_{\infty}^*(\boldsymbol{\mu}_0, \cdot)$ \citep{paternain2019constrained}, we obtain:
	\begin{align*}
		&V_{\infty}^*(\boldsymbol{\mu}_0, -2G_C)\geq \left(\dfrac{\zeta_1-2G_C}{\zeta_1-G_C}\right)V_{\infty}^*(\boldsymbol{\mu}_0, -G_C)+ \left(\dfrac{G_C}{\zeta_1-G_C}\right)V_{\infty}^*(\boldsymbol{\mu}_0, -\zeta_1),\\
		\text{Equivalently, }	&V_{\infty}^*(\boldsymbol{\mu}_0, -G_C) - V_{\infty}^*(\boldsymbol{\mu}_0, -2G_C) \leq \left(\dfrac{G_C}{\zeta_1-G_C}\right)\left[V_{\infty}^*(\boldsymbol{\mu}_0, -G_C) - V_{\infty}^*(\boldsymbol{\mu}_0, -\zeta_1)\right]\\
		&\hspace{4.5cm}\leq \left(\dfrac{2G_C}{\zeta_1-G_C}\right)\left(\dfrac{M_R}{1-\gamma}\right)\leq \left(\dfrac{2G_C}{\zeta_1}\right)\left(\dfrac{M_R}{1-\gamma}\right)
	\end{align*}
	
	Combining, we obtain,
	\begin{align}
		\label{eq_51}
		\begin{split}
			\Big|V_N^*(\boldsymbol{x}_0^N, 0) 	-\dfrac{1}{J}\sum_{j=1}^{J}V_{\infty}^R(\boldsymbol{\mu}_0, \pi_{\Phi_j})\Big|\leq C_1\dfrac{1}{\sqrt{J}} &+ C_2\left(\sqrt{\epsilon_{\mathrm{bias}}}+\dfrac{C_3}{\sqrt{L}}\right) +G_R \\
			&+ G_C\left[\dfrac{4}{\zeta_0}\left(\dfrac{M_R}{1-\gamma}\right)\right] + \left(\dfrac{2G_C}{\zeta_1}\right)\left(\dfrac{M_R}{1-\gamma}\right)
		\end{split}
	\end{align}
	
	From Theorem 3 of \citep{ding2020natural}, we also obtain the following.
	\begin{align}
		\label{eq_50}
		&\dfrac{1}{J}\sum_{j=1}^{J}V_{\infty}^C(\boldsymbol{\mu}_0, \pi_{\Phi_j})+2G_C\leq \dfrac{C_3}{J^{\frac{1}{4}}} +\dfrac{C_4}{J^{\frac{1}{4}}}\left(\epsilon_{\mathrm{bias}}^{\frac{1}{4}}+\dfrac{C_5}{L^{\frac{1}{4}}}\right)
	\end{align}
	for some constants $C_4, C_5$. Using Lemma \ref{lemma_1a}, we obtain,
	\begin{align}
		\label{eq_52}
		\dfrac{1}{J}\sum_{j=1}^JV_N^C(\boldsymbol{x}_0^N, \pi_{\Phi_j}) \leq \dfrac{1}{J}\sum_{j=1}^J V_\infty^C(\boldsymbol{\mu}_0, \pi_{\Phi_j}) + G_C \leq 	 \dfrac{C_3}{J^{\frac{1}{4}}} +\dfrac{C_4}{J^{\frac{1}{4}}}\left(\epsilon_{\mathrm{bias}}^{\frac{1}{4}}+\dfrac{C_5}{L^{\frac{1}{4}}}\right) -G_C
	\end{align}
	
	Note that, $G_R, G_C=\mathcal{O}(e)$ where $e\triangleq\left(\frac{1}{\sqrt{N}}\left[\sqrt{|\mathcal{X}|}+\sqrt{|\mathcal{U}|}\right]\right)$. To make the RHS of $(\ref{eq_52})$ negative, we must have $J=\mathcal{O}(G_C^{-4})=\mathcal{O}(e^{-4})$. Moreover, if we choose $L=\mathcal{O}(e^{-2})$, then the RHS of $(\ref{eq_49})$ becomes $\mathcal{O}(e+\sqrt{\epsilon_{\mathrm{bias}}})$. Hence, the desired sample complexity is $\mathcal{O}(e^{-6})$. As discussed earlier, this sample complexity might be improved to $\Tilde{\mathcal{O}}(e^{-2})$ if the recently proposed algorithm of \cite{mondal2024sample} is used.

	\newpage
	\section{Sampling Procedure}
	\label{sec:appndx_sampling}

 \begin{algorithm}[h!]
		\caption{Sampling Algorithm}
		\label{algo_2}
		\textbf{Input:} $\boldsymbol{\mu}_0$, $\boldsymbol{\pi}_{\Phi_j}$,
		$P$, $r$
		\begin{algorithmic}[1]
			\STATE Sample $x_0\sim \boldsymbol{\mu}_0$. 
			\STATE Sample $u_0\sim {\pi}_{\Phi_j}(x_0,\boldsymbol{\mu}_0)$ 
			\STATE  $\boldsymbol{\nu}_0\gets\nu^{\mathrm{MF}}(\boldsymbol{\mu}_0,\pi_{\Phi_j})$
			\vspace{0.2cm}
			\STATE $t\gets 0$ 
			\STATE $\mathrm{FLAG}\gets \mathrm{FALSE}$
			\WHILE{$\mathrm{FLAG~is~} \mathrm{FALSE}$}
			{
				\STATE $\mathrm{FLAG}\gets \mathrm{TRUE}$ with probability $1-\gamma$.
				\STATE Execute $\mathrm{Update}$
				
			}
			\ENDWHILE
			
			\STATE $T\gets t$
			
			\STATE Accept   $(x_T,\boldsymbol{\mu}_T,u_T)$ as a sample.

			\vspace{0.2cm}
			\STATE $\hat{V}_{\Phi_j}^R\gets 0$, $\hat{Q}^R_{\Phi_j}\gets 0$, $\hat{V}_{\Phi_j}^C\gets 0$, $\hat{Q}^C_{\Phi_j}\gets 0$
			
			\STATE $\mathrm{FLAG}\gets \mathrm{FALSE}$
			\STATE $\mathrm{SumRewards}\gets 0$
			\STATE $\mathrm{SumCosts}\gets 0$
			\WHILE{$\mathrm{FLAG~is~} \mathrm{FALSE}$}
			{
				\STATE $\mathrm{FLAG}\gets \mathrm{TRUE}$ with probability $1-\gamma$.
				\STATE Execute $\mathrm{Update}$
				\STATE $\mathrm{SumRewards}\gets \mathrm{SumRewards} + r(x_t,u_t,\boldsymbol{\mu}_t,\boldsymbol{\nu}_t)$
				\STATE $\mathrm{SumCosts}\gets \mathrm{SumCosts} + c(x_t,u_t,\boldsymbol{\mu}_t,\boldsymbol{\nu}_t)$
				
			}
			\ENDWHILE
			
			\vspace{0.2cm}
			\STATE With probability $\frac{1}{2}$, $\hat{V}_{\Phi_j}^R\gets \mathrm{SumRewards}$. Otherwise $\hat{Q}_{\Phi_j}^R\gets \mathrm{SumRewards}$.  
			\STATE With probability $\frac{1}{2}$, $\hat{V}_{\Phi_j}^C\gets \mathrm{SumCosts}$. Otherwise $\hat{Q}_{\Phi_j}^C\gets \mathrm{SumCosts}$.  
			
			\STATE $\hat{A}_{\Phi_j}^R(x_T,\boldsymbol{\mu}_T,u_T)\gets 2(\hat{Q}_{\Phi_j}^R-\hat{V}_{\Phi_j}^R)$.
			\STATE $\hat{A}_{\Phi_j}^C(x_T,\boldsymbol{\mu}_T,u_T)\gets 2(\hat{Q}_{\Phi_j}^C-\hat{V}_{\Phi_j}^C)$.
			\STATE $\hat{A}_{\Phi_j}^{\lambda_j}(x_T,\boldsymbol{\mu}_T,u_T)\gets\hat{A}_{\Phi_j}^R(x_T,\boldsymbol{\mu}_T,u_T)-\lambda_j \hat{A}_{\Phi_j}^C(x_T,\boldsymbol{\mu}_T,u_T) $
			
		\end{algorithmic} 
		\textbf{Output}: $(x_T,\boldsymbol{\mu}_T,u_T)$, $\hat{A}^{\lambda_j}_{\Phi_j}(x_T,\boldsymbol{\mu}_T,u_T)$, and $\hat{V}_{\Phi_j}^C$

		\vspace{0.3cm}
		
		\textbf{Procedure} $\mathrm{Update}$:
		
		\begin{algorithmic}[1]
			\STATE $x_{t+1}\sim P(x_t,u_t,\boldsymbol{\mu}_t,\boldsymbol{\nu}_t)$.
			\STATE  $\boldsymbol{\mu}_{t+1}\gets P^{\mathrm{MF}}(\boldsymbol{\mu}_t,\pi_{\Phi_j})$ 
			\STATE  $u_{t+1}\sim {\pi}_{\Phi_j}(x_{t+1},\boldsymbol{\mu}_{t+1})$
			\STATE $\boldsymbol{\nu}_{t+1}\gets\nu^{\mathrm{MF}}(\boldsymbol{\mu}_{t+1},\pi_{\Phi_j})$
			\STATE $t\gets t+1$
		\end{algorithmic}
		\textbf{EndProcedure}
  \vspace{0.2cm}
	\end{algorithm}

    \newpage
	\bibliography{Bib}

\end{document}